%% file: iclr2027_conference.tex
\documentclass{article} 
\usepackage{iclr2027_conference,times}

\input{math_commands.tex}

\usepackage{hyperref}
\usepackage{url}
\expandafter\def\expandafter\UrlBreaks\expandafter{\UrlBreaks\do\-}

\usepackage[utf8]{inputenc}
\usepackage[T1]{fontenc}
\usepackage{booktabs}
\usepackage{nicefrac}
\usepackage{microtype}
\usepackage{xcolor}
\usepackage{graphicx}
\usepackage{float}
\usepackage{subcaption}
\usepackage{array}
\usepackage{enumitem}
\usepackage{siunitx}
\usepackage{listings}
\usepackage{placeins}
\usepackage{wrapfig}
\usepackage{caption}
\usepackage{dblfloatfix}
\usepackage{afterpage}
\usepackage{amssymb}
\usepackage{mathtools}
\usepackage{amsthm}
\usepackage{multirow}
\usepackage[most]{tcolorbox}
\usepackage{colortbl}
\usepackage[capitalize,noabbrev]{cleveref}

\usepackage{algorithm}
\usepackage{algorithmic}

\definecolor{varBlue}{HTML}{276891}    
\definecolor{varOrange}{HTML}{E88645}  
\definecolor{varYellow}{HTML}{C9A03A}  
\definecolor{varSage}{HTML}{6F8E5A}    
\definecolor{varGreen}{HTML}{4F7A6E}   
\definecolor{varPink}{HTML}{B0568E}    

\definecolor{tintBlue}{HTML}{DCE6EE}
\definecolor{tintOrange}{HTML}{FBE2CD}
\definecolor{tintYellow}{HTML}{FAEDC4}
\definecolor{tintSage}{HTML}{E2EAD7}
\definecolor{tintGreen}{HTML}{D6E2DD}
\definecolor{tintPink}{HTML}{F1D8E4}

\newcommand{\hlname}[3]{%
  {\setlength{\fboxsep}{1.5pt}%
   \colorbox{#1}{\strut\textcolor{#2}{\texttt{\small #3}}}}%
}

\newcommand{\hlDL}[1]{\hlname{tintOrange}{varOrange!70!black}{#1}}
\newcommand{\hlDLC}[1]{\hlname{tintYellow}{varYellow!70!black}{#1}}
\newcommand{\hlDLM}[1]{\hlname{tintSage}{varSage!75!black}{#1}}
\newcommand{\hlGS}[1]{\hlname{tintGreen}{varGreen!75!black}{#1}}

\definecolor{jsonString}{RGB}{0,128,0}
\definecolor{jsonKeyword}{RGB}{160,0,0}
\definecolor{jsonBg}{RGB}{248,248,248}

\lstdefinelanguage{json}{
  basicstyle=\ttfamily\small,
  showstringspaces=false,
  breaklines=true,
  breakatwhitespace=false,
  columns=fullflexible,
  keepspaces=true,
  string=[b]",
  stringstyle=\color{jsonString},
  morekeywords={true,false,null},
  keywordstyle=\color{jsonKeyword}\bfseries
}

\lstdefinestyle{jsonIdle}{
  language=json,
  backgroundcolor=\color{jsonBg},
  frame=single,
  rulecolor=\color{black!20},
  numbers=left,
  numberstyle=\tiny\color{black!50},
  xleftmargin=2em,
  framexleftmargin=1.5em
}

\definecolor{pyKeyword}{RGB}{0,0,160}
\definecolor{pyComment}{RGB}{160,0,0}
\definecolor{pyString}{RGB}{0,128,0}

\lstdefinestyle{pyIdle}{
  language=Python,
  basicstyle=\ttfamily\small,
  showstringspaces=false,
  breaklines=true,
  breakatwhitespace=false,
  columns=fullflexible,
  keepspaces=true,
  backgroundcolor=\color{jsonBg},
  frame=single,
  rulecolor=\color{black!20},
  numbers=left,
  numberstyle=\tiny\color{black!50},
  xleftmargin=2em,
  framexleftmargin=1.5em,
  morestring=[s]{"""}{"""},
  morestring=[s]{'''}{'''},
  keywordstyle=\color{pyKeyword}\bfseries,
  commentstyle=\color{pyComment}\itshape,
  stringstyle=\color{pyString}
}

\newcommand{\safeincludegraphics}[2][]{%
  \IfFileExists{#2}{\includegraphics[#1]{#2}}{\fbox{Missing file: \texttt{#2}}}%
}

\theoremstyle{plain}

\theoremstyle{definition}

\theoremstyle{remark}

\title{An Investigation of Robustness of LLMs in Mathematical Reasoning:
  Benchmarking with Mathematically-Equivalent Transformation of Advanced Mathematical Problems}

\author{%
  Yuren Hao\\
  Department of Computer Science\\
  University of Illinois Urbana--Champaign\\
  Urbana, IL, USA\\
  \texttt{yurenh2@illinois.edu}\\
  \And
  Xiang Wan\\
  Department of Computer Science\\
  Stanford University\\
  Stanford, CA, USA\\
  \texttt{oscarwan@stanford.edu}\\
  \And
  Chengxiang Zhai\\
  Department of Computer Science\\
  University of Illinois Urbana--Champaign\\
  Urbana, IL, USA\\
  \texttt{czhai@illinois.edu}\\
}

\iclrpreprintcopy 
\begin{document}

\maketitle

\begin{abstract}
  \input{sections/00_abstract}
\end{abstract}

\section{Introduction}
\label{sec:introduction}
\input{sections/01_introduction}

\section{Related Work}
\label{sec:related_work}
\input{sections/02_related_work}

\section{The GAP Framework}
\label{sec:gap_framework}
\input{sections/03_gap_framework}

\section{PutnamGAP Instantiation}
\label{sec:putnamgap}
\input{sections/04_putnamgap}

\section{Failure Analysis}
\label{sec:evidence}
\input{sections/05_evidence}

\section{Discussion}
\label{sec:discussion}
\input{sections/06_discussion}

\subsection*{AI use statement}

LLMs are part of the method studied in this paper. \emph{As pipeline
components,} \textsc{o3} proposes surface-rename candidates and topical
metadata, executes the five kernel-generation stages, and serves as each of
the $J{=}5$ judges in the repair-and-verify loop (\S\ref{sec:pipelines},
Appendix~\ref{app:pipelines}); LLM graders score model solutions against
reference answers and rubrics, cross-checked across three providers and by
human audit (Appendix~\ref{app:grader_kappa}); \textsc{DeepSeek-V4-Flash}
serves as the outcome-blind difficulty judge and \textsc{Gemini-2.5-Flash} as
the backend for the non-Putnam feasibility check
(Appendices~\ref{app:difficulty} and~\ref{app:kernel_beyond_putnam}). All
prompts are listed in Appendix~\ref{app:prompts}. \emph{As evaluated
systems,} the 18 models of the main study, \textsc{DeepSeek-V4-Pro},
\textsc{DeepSeek-V4-Flash}, and the Intel DeepMath agent are the subjects of
evaluation.

Separately, we used generative AI tools (Claude and Codex, as coding and writing
agents) for writing and debugging experiment and analysis code, editing and
reorganising the manuscript, formatting tables and bibliography entries, and
searching for related work. All experimental designs, claims and conclusions
are the authors' own; every number reported was produced by our code and
checked by the authors, and every citation was checked against its source.
We have reviewed all AI-assisted work and take responsibility for the final
content of this work, including text, claims and artifacts produced with the
aid of generative AI.

\subsection*{Ethics statement}

PutnamGAP is derived from published Putnam problems and solutions. Items from 1938--2016 are digitised from the four MAA Press monographs under their fair-use clause for teaching and research, and post-2017 items are included with MAA's permission; we release machine-readable text and identifiers but not the scanned books (Appendix~\ref{app:sources}). The study involves no human subjects beyond author and volunteer quality checks of problem text, and no personal data.

\subsection*{Reproducibility statement}

The variant-generation pipelines, prompts, and evaluation methodology are described in
Section~\ref{sec:gap_framework} and Section~\ref{sec:putnamgap}, with full implementation
details in Appendices~\ref{app:pipelines}, \ref{app:algorithm}, and~\ref{app:prompts}.
Data sources and the OCR pipeline are documented in Appendix~\ref{app:sources}.
Ablation studies appear in Appendix~\ref{app:ablations}.
The complete PutnamGAP dataset and evaluation code are provided as supplementary material.

\bibliography{example_paper}
\bibliographystyle{iclr2027_conference}


\appendix
\raggedbottom

\input{sections/appendix/00_appendix}

\end{document}

%% file: math_commands.tex
\usepackage{amsmath,amsfonts,bm}

\def\eqref#1{equation~\ref{#1}}

\def\1{\bm{1}}

\DeclareMathAlphabet{\mathsfit}{\encodingdefault}{\sfdefault}{m}{sl}
\SetMathAlphabet{\mathsfit}{bold}{\encodingdefault}{\sfdefault}{bx}{n}



%% file: sections/00_abstract.tex
Frontier large language models (LLMs) now reach near-ceiling accuracy on standard mathematical-reasoning benchmarks and gold-medal-level performance at the International Mathematical Olympiad. As these benchmarks saturate and their items leak into training data, a high score no longer shows whether a model reasons robustly, nor which component of its reasoning fails when it does not. 
To evaluate reasoning in a way that remains informative for such models and makes its failures diagnosable, 
we propose a novel evaluation methodology
called \textbf{GAP}
(Generalisation--and--Perturbation), where the key idea is to automatically generate many mathematically-equivalent variants of an existing mathematics problem at scale, using two disjoint interpretable transformations: 
1) \emph{surface renames}, which probe the
binding between identifiers and latent variable roles, and 2)
\emph{kernel rewrites}, which probe whether the high-level proof plan
 survives a change of mathematical setting. Compared with existing benchmarks, a benchmark created using GAP has two important benefits: 1) Since the new variants generated are unlikely ``seen" by an LLM, a GAP benchmark effectively addresses the data leakage problem. 2) A GAP benchmark enables diagnosis of reasoning failures based on the performance of an LLM over a spectrum of variants of mathematics problems designed with different transformations, each corresponding to a meaningful hypothesis about the cause of failure. 
We instantiate GAP on all of the 1{,}051 William Lowell Putnam Competition problems from 1938 to 2024, adding 5{,}255 unseen transformed instances to form \textbf{PutnamGAP}, a 6{,}306-item competition-level mathematics corpus, resulting in  
the first public machine-readable dataset built from the \emph{full} Putnam archive. 
Using PutnamGAP, we evaluated 18 commercial and open-source models spanning different sizes and providers, and found that accuracy drops across all models and variant families, with the largest degradations under kernel rewrites. This gap does not close with model strength, suggesting that the dominant remaining weakness of even the strongest models is carrying a proof plan into a changed setting, not handling surface changes. Further diagnostic analysis enables more detailed understanding of the causes of failure and generates potentially useful insights for future research on improving the reasoning capacity of LLMs. 

%% file: sections/01_introduction.tex

Reasoning models that scale inference-time computation \citep{snell:2024, openai:2024o1, deepseek:2025r1} have turned mathematics from a weakness of large language models into one of their headline capabilities, from olympiad medals \citep{hubert:2025, deepmind:2025imo} to near-ceiling accuracy on long-standing benchmarks \citep{Cobbe:2021, hendrycks:2021}. The benchmarks built to measure this progress are being exhausted as quickly: static item sets saturate, their contents leak into training corpora \citep{sainz:2023, balloccu:2024}, and even harder successors \citep{glazer:2024, balunovic:2025} record how often a model is right rather than what it has learned.

A high score may reflect robust proof construction, memorized solutions, or successful exploitation of surface cues; a low score under a modified problem shows that something broke, but not what broke. This paper asks a more diagnostic question: when a model fails on a transformed mathematical problem that it could solve in canonical form, which component of its reasoning failed?

Perturbation benchmarks partially address this problem by testing whether models remain accurate after paraphrases, numerical changes, variable renames, or semantic modifications \citep{Li:2024GSMPlus, Mirzadeh:2024GSMSymbolic, huang:2025, gulati:2025}. These evaluations establish that brittleness exists. However, when different transformation types are mixed into a single robustness score, the resulting drop is hard to interpret: it may reflect lexical instability, contamination, semantic drift, increased difficulty, or failure to transfer a proof strategy. Heavy human curation can improve quality \citep{Li:2024GSMPlus, huang:2025}, but it does not by itself make the perturbation axis diagnostic.

To address these limitations, we propose a novel diagnostic evaluation method called \textbf{GAP}
(Generalisation--and--Perturbation) to deeply analyze the robustness of reasoning. 
We focus on two requirements that are elementary in mathematics but distinct for LLMs. The first is symbol-role binding: if a problem is unchanged except that variables and parameters receive unfamiliar, misleading, or uninformative names, a robust solver should preserve the binding between each identifier and its latent mathematical role. The second is proof-plan persistence: if the concrete setting and answer change but the high-level proof plan remains valid, a robust solver should transfer the plan rather than restart from surface similarity.

In this work, surface transformations are answer-preserving equivalences, whereas kernel transformations are proof-plan-preserving rewrites: they change the instance and answer while preserving the abstract sequence of mathematical moves.

GAP constructs paired variants of the same source problem. Surface renames leave the mathematical statement, answer, and reference proof path unchanged while changing the identifiers along a spectrum from descriptive to misleading or garbled. Kernel rewrites change the mathematical setting and terminal answer while preserving the high-level proof plan. Because both families are applied to the same canonical items, a model’s behavior can be compared within item rather than only across unrelated benchmark examples. The goal is not only to measure whether accuracy drops, but to make the source of the drop experimentally separable.

We instantiate GAP on the full William Lowell Putnam Competition archive from 1938 to 2024 \citep{putnamI, putnamII, putnamIII, putnamIV}, yielding PutnamGAP: 1,051 canonical problems and 5,255 aligned transformed instances. This setting is useful for three reasons. First, Putnam problems are advanced, proof-heavy, and require multi-step reasoning rather than only computation. Second, the aligned construction lets us compare original, surface, and kernel behavior on the same mathematical source item. Third, the transformed variants reduce reliance on static benchmark memorization and allow fresh instances to be generated prospectively. The surface variants are deterministic renames, while kernel variants are produced by a robust proof-DAG pipeline.

Across 18 commercial and open-source models, accuracy drops under all transformed families, with the largest losses under kernel rewrites, and the kernel--surface gap remains large for the strongest models and on a newer model (Appendix~\ref{app:newer_model}). However, the central finding is not merely that kernel variants are harder. The two transformation families produce qualitatively different failure signatures. Surface failures drift from the model’s own original-correct trajectory, suggesting a form of self-inconsistency under changed identifiers. Kernel failures drift from the canonical proof plan itself, suggesting failure to preserve or transfer the intended strategy. A prefix-restoration intervention further separates the regimes: an expert-solution prefix substantially improves recovery on surface failures but has limited effect on kernel failures. Thus, mathematical robustness is not a scalar property: two models, or two perturbation families, can exhibit similar aggregate drops while failing for different reasons and responding differently to intervention.

Our contributions are:
\begin{enumerate}[nosep,leftmargin=1.4em]
\item We propose GAP, a mechanism-aligned perturbation framework that separates answer-preserving surface renames from proof-plan-preserving kernel rewrites.
\item We instantiate GAP as PutnamGAP, a 6,306-item aligned corpus built from the full Putnam archive, with deterministic surface generation and verified kernel generation.
\item We evaluate 18 LLMs and show that kernel rewrites cause substantially larger capability losses than surface renames, even when conditioning on items solved in canonical form, and that this gap does not close for the strongest models.
\item We provide trajectory-overlap and prefix-restoration analyses showing that surface and kernel failures correspond to different observable regimes: repairable self-inconsistency versus plan drift resistant to the same intervention.
\end{enumerate}

%% file: sections/02_related_work.tex

\paragraph{Competition-level mathematical reasoning.}
Benchmarks at Olympiad or research level include MATH \citep{hendrycks:2021}, OlympiadBench \citep{he:2024}, Omni-MATH \citep{gao:2024}, PutnamBench \citep{tsoukalas:2024}, FrontierMath \citep{glazer:2024}, MathArena \citep{balunovic:2025}, LiveAoPSBench \citep{mahdavi:2025}, and IMO-Bench \citep{luong:2025imobench}, whose problems are altered by experts to limit memorisation. These establish difficulty ceilings but do not control the perturbation axis, so they measure capability but not robustness; answer-only scores can also overstate reasoning, as models near human level on AIME answers scored below $25\%$ on USAMO~2025 proofs \citep{petrov:2025proof}.

\paragraph{Automated perturbation benchmarks.}
GSM-Symbolic \citep{Mirzadeh:2024GSMSymbolic} uses templates to resample numerical values; GSM-Plus \citep{Li:2024GSMPlus} and MATH-Perturb \citep{huang:2025} add human-curated semantic and structural modifications; Putnam-AXIOM \citep{gulati:2025} applies programmatic rewrites to a subset of Putnam problems; MaSTer \citep{hou:2025} runs a multi-round LLM rewrite--verify loop to stress-test targets on GSM8K and MATH-500; RRB \citep{golikov:2026rrb} applies 13 deterministic textual perturbations to AIME problems and finds frontier models largely resilient; and Math-RoB \citep{yu:2025} synthesises instruction-level modifications. All demonstrate that brittleness exists, but because they mix transformation types into a single robustness score, the resulting drop cannot be attributed to a specific failure mechanism.

\paragraph{Perturbation-axis decomposition for mechanism analysis.}
A smaller line of work aims to decompose \emph{why} models fail. \citet{stolfo:2022} introduce a causal framework that quantifies robustness of arithmetic reasoning under controlled variable swaps. Fragile Reasoning \citep{han:2026} applies meaning-preserving surface perturbations to GSM8K and probes internal activations to localise failure. \citet{essebbani:2026} construct parameterised 2-SAT instances that separate surface from structural difficulty in the logic domain. MATH-Perturb \citep{huang:2025} splits perturbations into ``easy'' (surface) and ``hard'' (structural) bins, though the hard variants break equivalence and therefore confound difficulty with transformation type. \citet{yang:2025numerical} classify failures under numerical variation as calculation or reasoning errors after the fact; GAP instead fixes the tested invariance before evaluation (Appendix~\ref{app:related_extended}).\looseness=-1

No prior work fully satisfies even two of these three desiderata: Putnam-AXIOM is automated at competition level but does not decompose the perturbation axis; MATH-Perturb decomposes at competition level but its hard variants break equivalence; Fragile Reasoning automates and decomposes but operates on GSM-level problems. GAP is the first framework to satisfy all three simultaneously.

%% file: sections/03_gap_framework.tex
%

\subsection{Two transformation families, one diagnostic split}
\label{sec:two-families}

GAP applies \emph{equivalence-preserving} transformations $\tau$ to a
canonical item $(x,y,\pi)$ -- problem statement~$x$, reference
answer~$y$, and an unreleased expert solution path~$\pi$ that the
pipeline uses internally. A model $f_\theta$ that consumes $\tau(x)$ is
graded by the same automatic checker as on $x$, yielding a binary label
$z=\operatorname{grade}(f_\theta(\tau(x)),y_\tau)\in\{0,1\}$, where
$y_\tau$ tracks any change of answer induced by $\tau$. We then partition
$\tau$ into two disjoint super-families:
\begin{itemize}[nosep,leftmargin=1.2em]
  \item $\mathcal{T}^{\mathrm{surf}}$: bijective renames of the
        problem's identifiers -- the free variables (e.g.\ $r,h$) and
        the fixed parameters introduced by the statement
        (e.g.\ a given surface area $S$). Mathematical constants such
        as $\pi$ and $e$ are left untouched, and the statement length,
        equation structure, and reference solution path are preserved.
        We instantiate four sub-families that span the lexical-cue
        spectrum from \emph{helpful} to \emph{no signal} (\textsc{DL},
        \textsc{DLC}, \textsc{DLM}, \textsc{GS}; defined in
        Figure~\ref{fig:variant_example}).
  \item $\mathcal{T}^{\mathrm{kern}}$: rewrites that resample numerical
        and relational \emph{slots} of the problem while preserving the
        high-level proof plan. Both the statement and the answer change;
        the abstract sequence of mathematical moves does not. We
        instantiate one variant per source item, denoted \textsc{KV}.
\end{itemize}

This split is intentional, not cosmetic. Surface variants stress one
specific failure mode -- whether the model preserves a stable binding
between unfamiliar names and the latent variable roles they denote --
while keeping the proof plan untouched. Kernel variants stress the
complementary mode -- whether the model preserves the high-level plan
when the mathematical setting moves -- while keeping the local algebra
type unchanged. With both families running on the same item, a drop on
$\mathcal{T}^{\mathrm{surf}}$ but not $\mathcal{T}^{\mathrm{kern}}$
implicates representation, the converse implicates planning, and a drop
on both points to a more global brittleness. The mechanism analysis in
Section~\ref{sec:evidence} tests exactly this distinction;
without the decomposition the question cannot even be posed.

\input{sections/fig_variant_example}

\paragraph{Evaluation matrix.}
For a corpus of $N$ canonical items and $K=K_{\mathrm{surf}}+K_{\mathrm{kern}}$
variants per item, running $f_\theta$ on every $(x_i,\tau)$ pair fills a
binary matrix $\mathbf{Z}\!\in\!\{0,1\}^{N\times(K+1)}$. The original
column gives the per-item correctness vector $e\!\in\!\{0,1\}^N$, and
the family-wise aggregates $h^{\mathrm{surf}}\!=\!\operatorname{maj}(\mathbf{Z}_{[:,\mathrm{surf}]})$
and $h^{\mathrm{kern}}\!=\!\mathbf{Z}_{[:,\mathrm{kern}]}$ form the
substrate for the paired analyses in
Section~\ref{sec:evidence}: per-item accuracy drops, paired
significance tests, and the structural-overlap and rescue
contrasts that decompose those drops by mechanism.\looseness=-1

\subsection{Automated generation pipelines}
\label{sec:pipelines}

The two families are produced by two pipelines that share a quality
gate but differ sharply in cost. Both are end-to-end automatable: GAP
does not require the human curation step that, for instance,
\textsc{GSM-Plus} and \textsc{MATH-Perturb} use to refine their
candidate variants.

\paragraph{Surface pipeline (deterministic, cheap).}
Each free-variable or fixed-parameter token in the problem is sent to
a single \textsc{o3} call, conditioned on its role and its surrounding
context, which returns a candidate replacement string.
A deterministic post-validator rejects names that collide with any
other identifier already present in the problem; the accepted string is
then routed to one of the four family bins (\textsc{DL}, \textsc{DLC},
\textsc{DLM}, \textsc{GS}) by a lightweight tagger. Because the
operation is a pure rename, no theorem-proving capability is required
and no judge is needed: every accepted variant is mathematically
equivalent by construction. The full corpus of 1{,}051 items is
processed in roughly fifteen minutes of wall-clock time on a CPU-only
node with exponential back-off retries on transient API errors.

\begin{figure}[t]
  \centering
  \includegraphics[width=0.85\linewidth]{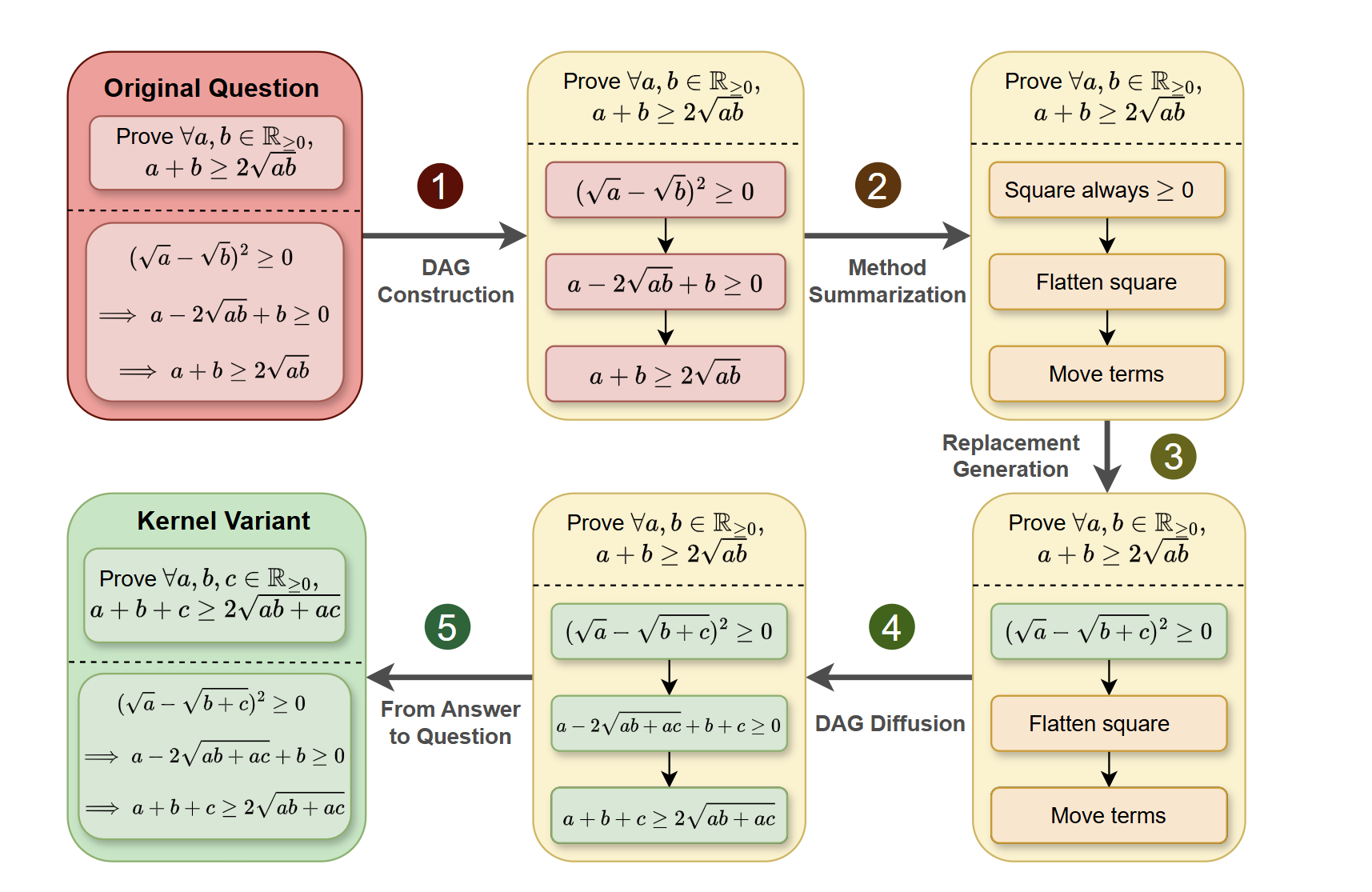}
  \caption{Kernel-variant pipeline, illustrated on arithmetic–geometric mean inequality. The proof DAG
  built from the canonical solution (stage~1) is abstracted into method
  labels (stage~2); a replacement is generated at the leaf
  (stage~3) and propagated through the DAG (stage~4); the modified DAG
  is rendered back into a question statement (stage~5). The plan is
  preserved by construction, so kernel variants stress \emph{plan
  persistence} under setting change rather than local algebra
  (Appendix~\ref{app:kern_invariant}).}
  \label{fig:kernel_pipeline}
\end{figure}

\paragraph{Kernel pipeline (multi-judge, structured).}
Kernel rewrites are harder because the statement, the answer, and most
of the algebra change while the proof plan must not. We therefore
factor the rewrite through an intermediate \emph{proof DAG} so that the
plan can be preserved by construction rather than by post-hoc
verification. The five stages are illustrated in
Figure~\ref{fig:kernel_pipeline} and proceed as follows.

\begin{enumerate}[nosep,leftmargin=1.4em]
\item \textbf{DAG construction.} The official solution $\pi$ is parsed
      into a directed acyclic graph whose nodes are intermediate claims
      and whose edges are local entailments.
\item \textbf{Method summarisation.} Each node is abstracted from a
      concrete equation into a content-free \emph{method label} (e.g.\
      ``square always $\geq 0$''; ``flatten square''; ``move terms''),
      yielding a plan that a human could re-execute on different inputs.
\item \textbf{Replacement generation.} A new instance of the leaf math
      object is sampled subject to a guard range derived from the
      problem's own inequalities (e.g.\ replacing $a+b\geq 2\sqrt{ab}$
      with $a+b+c\geq 2\sqrt{ab+ac}$).
\item \textbf{DAG diffusion.} The replacement is propagated through the
      dependent nodes by re-instantiating each method label with the
      new operands, yielding a regenerated proof whose step order
      matches the abstracted plan exactly.
\item \textbf{From answer to question.} The terminal node is rendered
      back into a fully natural-language problem statement, completing
      the variant.
\end{enumerate}

A repair-and-verify loop sits on top of stages 3--5. Each candidate is
shown to $J=5$ independent \textsc{o3} judges; the candidate is
\emph{accepted} only when two consecutive iterations both produce
unanimous \texttt{accept} verdicts (the ``two-in-a-row'' rule). On
disagreement, the suggested patch is applied and the candidate
re-enters the loop, up to $T=15$ iterations. Under an empirical
single-judge error rate $\varepsilon\!\le\!0.04$, a union bound over the
$T-K+1=14$ length-2 windows gives a soundness bound
$\delta\le 14\,\varepsilon^{KJ}<10^{-10}$ on a flawed variant slipping
through (Appendix~\ref{app:delta_soundness}). The protocol survives even at $\varepsilon=0.08$,
where $\delta<10^{-8}$. A 10\,\% manual audit on accepted variants is
applied in addition. Wall-clock time for the full corpus is under three
hours on commodity hardware.

\subsection{Reliability through information-asymmetric verification}
\label{sec:reliability}

A natural concern is that the kernel pipeline relies on an LLM both to
generate and (through judges) to verify candidate variants. The key
observation is that every LLM call in the pipeline performs
\emph{verification given a candidate} --- checking a presented proof
against a known answer --- rather than de novo problem solving. This
information asymmetry makes the task strictly easier than the original
problem, so the pipeline's reliability is \emph{architectural} rather
than dependent on the absolute capability of any single model.
Appendix~\ref{app:robustness} develops this argument in full, deriving
a soundness bound, stress-testing it under elevated error rates, and
cross-checking with independent grader providers and human audits.

%% file: sections/fig_variant_example.tex
\begin{figure}[t]
  \centering
  \footnotesize
  \setlength{\tabcolsep}{4pt}
  \renewcommand{\arraystretch}{1.55}
  \begin{tabular}{@{}p{0.085\linewidth}@{\hspace{4pt}}p{0.86\linewidth}@{}}
    \rowcolor{tintBlue!60}
    \textbf{\textcolor{varBlue}{Original}} &
      $V \;=\; \tfrac{5}{3}\,\pi\,r^{2}\,h$, \quad with $r,h$ free and
      surface area $S$ fixed. \\
    \rowcolor{tintOrange!50}
    \textbf{\textcolor{varOrange}{DL}} &
      $\hlDL{volume} \;=\; \tfrac{5}{3}\,\pi\,\hlDL{radius}^{2}\,\hlDL{altitude}$
      \quad{\scriptsize\textcolor{black!55}{(descriptive English; sanity
      check)}} \\
    \rowcolor{tintYellow!50}
    \textbf{\textcolor{varYellow}{DLC}} &
      $\hlDLC{calendar} \;=\; \tfrac{5}{3}\,\pi\,\hlDLC{monument}^{2}\,\hlDLC{daybreak}$
      \quad{\scriptsize\textcolor{black!55}{(unrelated nouns; semantic
      noise)}} \\
    \rowcolor{tintSage!55}
    \textbf{\textcolor{varSage}{DLM}} &
      $\hlDLM{surfacearea} \;=\; \tfrac{5}{3}\,\pi\,\hlDLM{perimeterlength}^{2}\,\hlDLM{depthvalue}$
      \quad{\scriptsize\textcolor{black!55}{(misleading: the new
      ``$V$'' is named \texttt{surfacearea})}} \\
    \rowcolor{tintGreen!55}
    \textbf{\textcolor{varGreen}{GS}} &
      $\hlGS{mnbvcxza} \;=\; \tfrac{5}{3}\,\pi\,\hlGS{qzxwvtnp}^{2}\,\hlGS{yrklsfhd}$
      \quad{\scriptsize\textcolor{black!55}{(garbled hashes; no lexical
      cue)}} \\
  \end{tabular}

  \vspace{2pt}

  \begin{tcolorbox}[
    enhanced, boxrule=0.4pt, arc=2pt, left=6pt, right=6pt,
    top=4pt, bottom=4pt,
    colback=tintPink!55, colframe=varPink!60!black,
    fontupper=\footnotesize,
    title={\textbf{\textcolor{white}{Kernel variant (\textsc{KV}): same
      plan, different setting}}},
    coltitle=white, colbacktitle=varPink!75!black,
    titlerule=0pt, fonttitle=\footnotesize,
  ]
    The two cones become \emph{one} cone of altitude $h/2$, and the
    coated area excludes the internal interface. The proof plan -- (i)
    write surface and volume in $r,h$, (ii) eliminate one variable from
    the constraint, (iii) differentiate, (iv) verify endpoints, (v)
    return the optimal ratio -- is unchanged. The terminal equation is
    no longer $h_0/r_0 = \tfrac{2}{5}\sqrt{5}$ but rather
    \[
    15\,(h/r)^{3} - 32\,(h/r)^{2} + 96\,(h/r) - 128 \;=\; 0,
    \]
    whose unique positive root is $h/r\approx 1.552$.
  \end{tcolorbox}

  \caption{One Putnam item (\texttt{1938-A-2}, the buoy problem) under
  each of GAP's five variants. The four surface families share the
  problem's algebra and reference proof; only the names of the free
  variables and fixed parameters change. The kernel variant changes
  the setting (number of cones, painted-area convention) and
  therefore the answer, but follows the same five-step proof plan as
  the original. Surface variants stress \emph{representation}; the
  kernel variant stresses \emph{plan persistence}.}
  \label{fig:variant_example}
\end{figure}

%% file: sections/04_putnamgap.tex
%

\subsection{Sources and contamination resistance}

PutnamGAP is built on every problem of the William Lowell Putnam
Mathematical Competition from 1938 to 2024, totalling \textbf{1{,}051}
distinct items after deduplication. The vast majority of pre-2017
problems are sourced from the four MAA Press monographs (Putnam I--IV),
which are distributed only as scanned print volumes; we obtain
\LaTeX\ for each item via a three-stage routine (manual problem
segmentation, MathPix-based formula-aware OCR, and custom post-filters
that fix the residual symbol errors), then spot-check every item
manually before any variant is generated. The precise breakdown
between scanned-only sources and post-2017 problems released under MAA
licence is given in Appendix~\ref{app:sources}; the practical consequence is that for
the bulk of the corpus, faithful machine-readable copies of either the
problem or its solution have not been part of any standard web crawl,
which limits the surface area for verbatim leakage into pre-training.

GAP adds a second, prospective layer of contamination resistance. Even
on the post-2017 fraction where leakage is more plausible, the kernel
pipeline can resample arbitrarily many fresh items \emph{after} a model
is trained: each of the 1{,}051 source items admits a bounded but
non-trivial number of equivalence-preserving rewrites, so
benchmark--training contamination is a generative pipeline problem
rather than a one-shot data problem. We use this property explicitly
in Section~\ref{sec:evidence}, where the mechanism analyses are validated on
freshly-generated kernel variants whose statements never existed at the
time the evaluated models were trained.

\subsection{Scale, coverage, and quality control}

\paragraph{Scale.}
Applying the four surface transformations and the kernel transformation
to each original yields $1{,}051\times(1+4+1)=\mathbf{6{,}306}$
aligned items. Part distribution is balanced (\textbf{527} from the A
half, \textbf{524} from the B half), and the canonical positional
identifier $\langle\textit{year},\textit{part},\textit{index}\rangle$
serves as a difficulty proxy: indices 1--2 are taken as Easy
($32.3\,\%$), 3--4 Medium ($32.3\,\%$), 5--6 Hard ($32.2\,\%$), with a
$3.0\,\%$ tail at indices 7--8 from the early competition years.

\paragraph{Topic and form coverage.}
Multi-tag topical labels in \texttt{\_meta.tag} confirm broad
mathematical coverage: \textbf{Algebra} (641 items),
\textbf{Analysis} (521), \textbf{Number Theory} (392),
\textbf{Combinatorics} (286), and \textbf{Geometry} (239). By form,
the corpus is dominated by proofs (\textbf{803} items) with the
remainder being calculations (\textbf{248}); both forms are graded
under the dedicated proof and computation rubrics detailed in
Appendix~\ref{app:prompts}.

\paragraph{Quality control.}
Surface variants are verified deterministically by the post-validator
described in Section~\ref{sec:pipelines} and require no judge.
Kernel variants pass the multi-judge two-in-a-row protocol with
$J{=}5$ and $T{\le}15$; no manual curation is part of the generation
pipeline. A separate $10\,\%$ post-hoc audit on accepted items found
no substantive errors. A
three-provider grader cross-check (Cohen's $\kappa\!=\!0.96$--$1.00$;
Appendix~\ref{app:grader_kappa}) confirms that the binary correctness labels used
throughout the rest of the paper are stable
across grader families. Full pipeline configuration, ablations, and
audit logs are deferred to Appendices~\ref{app:pipelines}, \ref{app:ablations}, and~\ref{app:trajectory}.\looseness=-1

\subsection{The headline accuracy picture}
\label{sec:headline}

We evaluated 18 commercial and open-source models on PutnamGAP under a
unified zero-shot prompt (full prompts in Appendix~\ref{app:prompts}).
Figure~\ref{fig:accuracy_by_variant} reports
each model's accuracy on the original and on each of the five
transformations, with models sorted top-to-bottom by accuracy on the
original split.

\begin{figure}[t]
  \centering
  \includegraphics[width=1.0\linewidth]{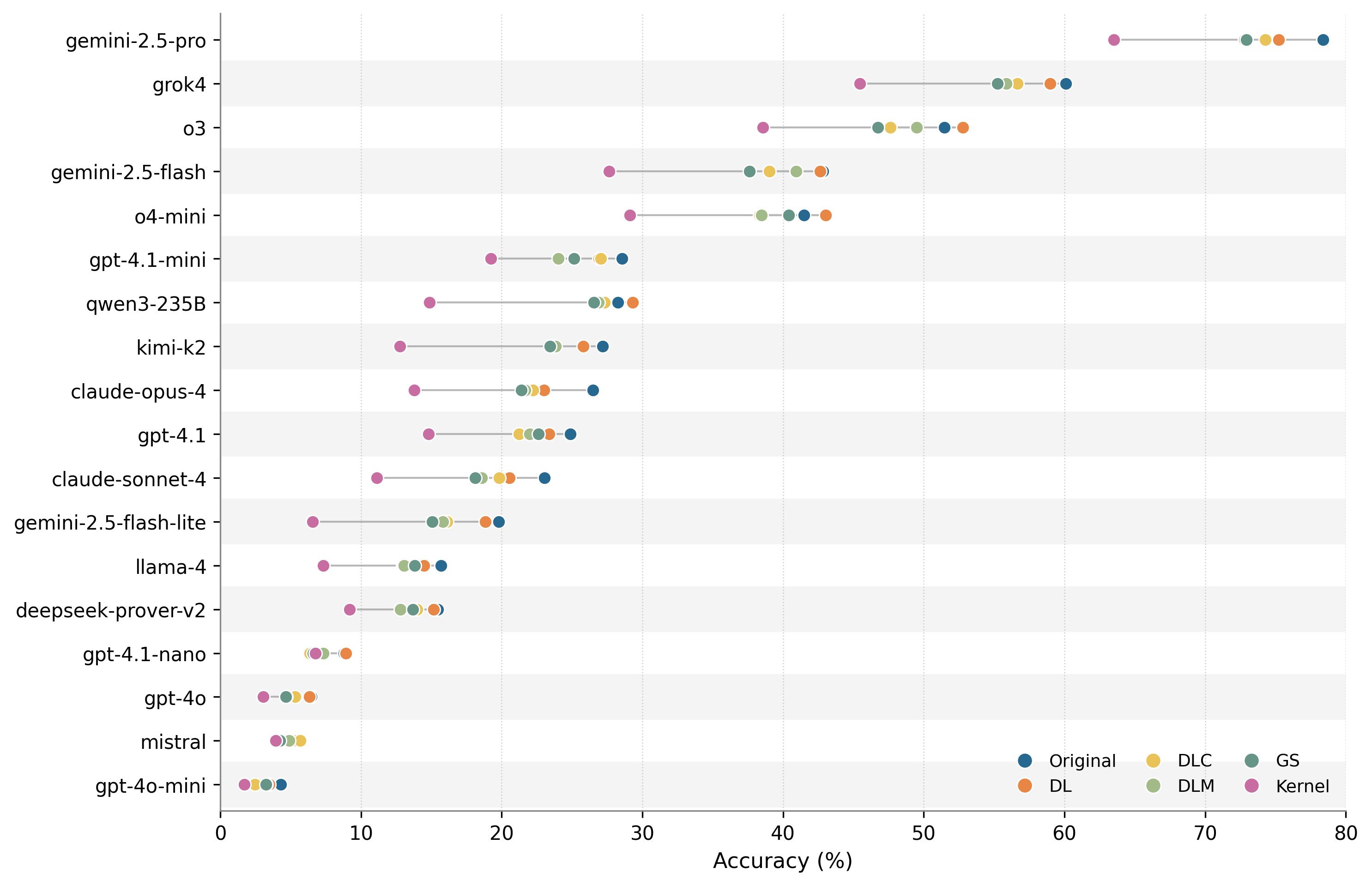}
  \caption{Accuracy by variant for the 18 evaluated models, sorted by
  performance on the original split. Each model's six variant scores
  are connected by a thin segment so the per-model trajectory is
  visible at a glance. Two patterns are nearly universal: (i) the
  surface variants form a tight cluster, with \textsc{DLC},
  \textsc{DLM}, and \textsc{GS} all sitting marginally below the
  original and \textsc{DL} essentially co-located with it; (ii) the
  kernel variant lies clearly to the left of the surface cluster for
  every model with non-trivial baseline accuracy, and the gap between
  the surface cluster and the kernel point remains large even for the
  strongest models. This pattern is the empirical
  phenomenon that Section~\ref{sec:evidence} dissects.}
  \label{fig:accuracy_by_variant}
\end{figure}

Two regularities are visible without any further analysis. First, the
surface variants behave as a tight cluster: \textsc{DL} accuracy is
within noise of the original, while \textsc{DLC}, \textsc{DLM}, and
\textsc{GS} land slightly below it -- the model loses a few points
when the rename is uninformative or actively misleading, but its
overall problem-solving capability is largely preserved. Second, the
kernel variant lies meaningfully below the surface cluster for nearly
every model: \texttt{o3} drops from $51.5\,\%$ on the original to
$38.6\,\%$ on \textsc{KV} (a $12.9$-point drop),
\texttt{gemini-2.5-pro} from $78.4\,\%$ to $63.5\,\%$ ($14.9$ points),
\texttt{grok4} from $60.1\,\%$ to $45.5\,\%$ ($14.6$ points), and the
gap survives across topic, difficulty, and form (Appendix~\ref{app:supplementary}).
Importantly, the kernel--surface gap does not close with model
strength: the five strongest models still lose $10$--$12$ points more
on kernel rewrites than on the average surface family.

These two patterns -- a flat surface cluster paired with a much larger
kernel drop -- are not by themselves evidence about \emph{why}
performance falls. They are the empirical handle on which the
mechanism analysis in Section~\ref{sec:evidence} is built: the surface/kernel
decomposition lets us ask whether the two drops have the same internal
signature, and the answer turns out to be no.

%% file: sections/05_evidence.tex
%
\subsection{Kernel variants break more solvable items than surface variants}

For each model we define the capability-loss rate $\mathcal{L} = b/(b{+}s)$, where $b$ counts items the model solves on the original but misses on the variant and $s$ counts items it solves on both. Unlike   
the raw accuracy delta, $\mathcal{L}$ conditions on items the model can actually solve, so it is not compressed by a low baseline. Across all 18 evaluated models, $\mathcal{L}^{\text{kern}}$ exceeds the mean
of $\mathcal{L}^{\text{surf}}$ over the four surface families (sign test, $p = 7.6 \times 10^{-6}$). The median $\mathcal{L}^{\text{kern}}$ is $70\%$; the median $\mathcal{L}^{\text{surf}}$ is $40\%$. In raw   
accuracy terms the median kernel drop is $-10.3$~pp versus $-2.1$~pp for the strongest surface family (the per-model breakdown behind Figure~\ref{fig:accuracy_by_variant} is tabulated with Wilson CIs in
Appendix~\ref{app:aggregate}). The asymmetry quantifies how much more kernel transformations break, but not why. Section~\ref{sec:dissociation} moves to the trajectory level, where the two failure modes dissociate.

\subsection{Surface and kernel failures dissociate at the trajectory level}
\label{sec:dissociation}

We compute token-Jaccard Cohen's $d$ (stable vs brittle) under two reference frames: self-anchor (the model's own original-correct trajectory, rename-mapped) and canonical anchor (the dataset's expert variant
   solution). Under self-anchor, all 72 surface cells (18 models $\times$ 4 variants) are direction-positive, with mean $d = {+}0.75$ and 69/72 reaching $p < 0.05$. Switching to canonical anchor collapses the
  surface effect to mean $d = {+}0.23$; for kernel variants, canonical anchor yields mean $d = {+}0.35$ across 17/18 models. The within-cell difference $\Delta d = d_{\text{self}} - d_{\text{canonical}}$ has   
  mean ${+}0.52$ (95\% CI $[{+}0.47,\, {+}0.58]$, cell-level bootstrap), positive in 71/72 cells.

To separate the overlap signal from trajectory length, we stratify each cell into variant-length tertiles and
  recompute $d$: the shortest tertile retains 72/72 direction-positive cells at mean $d = {+}0.84$, ruling out length as the driver. A random-pairing noise floor (median Jaccard $0.22$) sits well below the
  same-problem stable median ($0.48$), confirming the metric captures problem-specific reasoning structure. We label the two signatures self-inconsistency (surface trajectories drift from the model's own
  solution) and strategy drift (kernel trajectories drift from the canonical plan); these are names for the observable pattern, not claims about internal state.

\begin{figure}[t]
  \centering
  \includegraphics[width=0.9\linewidth]{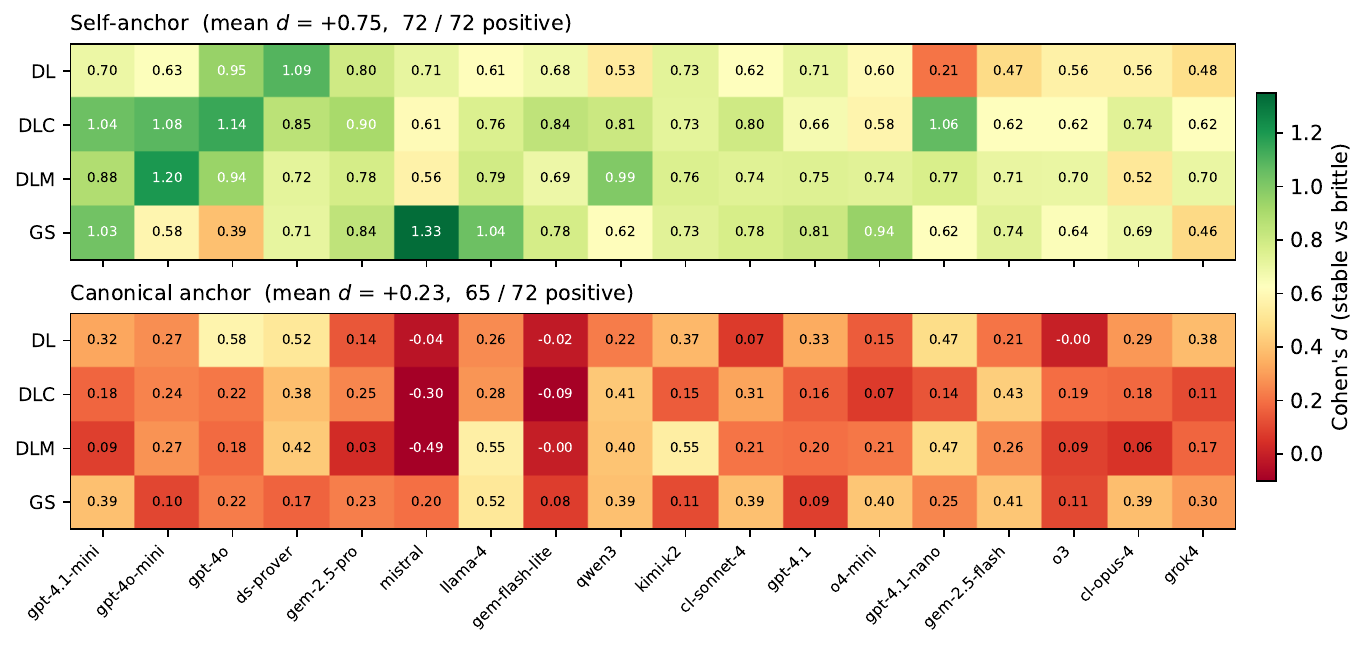}
  \caption{Structural-overlap Cohen's $d$ (stable vs brittle) under two reference frames. Each column is one model; each row is one surface variant. The self-anchor effect (top) collapses under canonical anchor (bottom).}
  \label{fig:heatmap_dissociation}
\end{figure}

\subsection{Expert-solution prefixes rescue surface failures but not kernel failures}

When a model fails a variant of a problem it solved in original form, does giving it the beginning of a correct solution help it recover? We re-prompt each such flip case under three conditions: null (variant problem only), canonical\_T2 (prefixed with the dataset's expert solution truncated at the first equation, after a leakage filter), and own\_T2 (prefixed with the model's own original-correct trajectory, rename-mapped and identically truncated). Pooling four models across the four surface families, canonical\_T2 raises the rebound rate significantly above null in every family (paired McNemar $b{=}90$, $c{=}28$, $\mathrm{OR} = 3.21$, $p = 8.9 \times 10^{-9}$); on kernel variants the same intervention does not reach significance.\looseness=-1

To isolate the contribution of plan content from prefix presence, we compare canonical\_T2 against own\_T2: both supply a partial-solution prefix in variant vocabulary, but only canonical\_T2 carries the expert plan. Canonical\_T2 wins the paired contrast ($b{=}75$, $c{=}36$, $p = 2.7 \times 10^{-4}$); own\_T2 is indistinguishable from null ($p = 0.08$).

A qualitative audit of 12 paired cases (Appendix~\ref{app:rescue_audit}) identifies five recurring archetypes of what canonical prefixes supply that own prefixes lack --- most commonly a closed-form ansatz or structural commitment that reframes the problem from discovery to verification. Even so, $55$--$60\%$ of surface flips remain unrescued under canonical\_T2, so the structural commitment is necessary but not sufficient.



%% file: sections/06_discussion.tex

\subsection{Implication}
The structural dichotomy identifies two failure modes but not their internal substrates; combining GAP's decomposition with model-internal probing of open-weight models \citep{meng:2022, belrose:2023, turner:2023} could localise each mode to specific layers or heads. The rescue asymmetry also suggests that inference-time interventions are mode-specific: an expert-solution prefix recovers many surface failures but few kernel failures, so kernel-side interventions remain an open problem. Finally, behavioural transfer to MATH \citep{hendrycks:2021} and ALG514 \citep{kushman-etal-2014-learning} (Appendix~\ref{app:cross_dataset}) shows that the pipeline generalises, but the trajectory-level dichotomy has so far been established only on PutnamGAP.\looseness=-1

\subsection{Conclusion}
Our paper tries to convey two messages. The first is methodological: equivalence-preserving transformations of mathematical problems can be cleanly decomposed into two disjoint, mechanism-aligned families (surface renames and kernel rewrites), and the resulting construction is generative, so the same pipeline can be re-applied after any future model release to produce fresh, contamination-resistant items and re-audit which components of reasoning are preserved. We realise this construction as \textbf{GAP}, and \textsc{PutnamGAP}, the benchmark we publish, is one instance of GAP applied to the full Putnam archive; the methodology itself is the contribution, and we expect GAP to remain a useful instrument for examining mathematical reasoning in models that do not yet exist. The second is empirical: the dominant remaining weakness in current LLMs' advanced mathematical reasoning is not surface fragility but an inability to carry a high-level proof plan through a change of mathematical setting. Across 18 commercial and open-source models evaluated on \textsc{PutnamGAP}, accuracy falls under every transformed family, but the loss is dominated by kernel rewrites, and the kernel--surface gap does not close for the strongest models. This, rather than further compression of already-small surface error, is the natural focal point for the next round of work on mathematical reasoning; Appendix~\ref{app:recommendations} turns it into concrete recommendations for evaluation and model development.

\subsection{Limitations}
We evaluate models under a single zero-shot pass (no test-time scaling); the $J{=}5$ kernel judges are drawn from a single model family; the structural diagnostic requires visible reasoning chains; \textsc{PutnamGAP} is English-only and competition-style; and generating and evaluating the full corpus incurs non-trivial API cost. Natural extensions are heterogeneous verifier ensembles that pair LLMs with symbolic checkers (Lean, SMT), porting the two-pipeline construction to adjacent STEM corpora with different proof structures, and applying surface and kernel transformations during fine-tuning to test whether the relevant invariances are learnable rather than incidental.

%% file: sections/appendix/00_appendix.tex
%
%

\input{sections/appendix/j_robustness}
\input{sections/appendix/k_additional_evaluations}
\input{sections/appendix/c_supplementary}
\input{sections/appendix/i_trajectory}
\input{sections/appendix/a_pipelines}
\input{sections/appendix/b_algorithm}
\input{sections/appendix/f_prompts}
\input{sections/appendix/d_sources}
\input{sections/appendix/e_ablations}
\input{sections/appendix/h_case_studies}
\input{sections/appendix/g_dataset_example}
\input{sections/appendix/l_related_work}
\input{sections/appendix/m_recommendations}

%% file: sections/appendix/j_robustness.tex
\section{Pipeline robustness analysis}
\label{app:robustness}

\subsection{Robustness of surface variations}

Surface variations are produced by deterministic string substitution from the dataset's rename map. A post-validator rejects any collision with existing identifiers. Because the operation is a pure rename --- no equation, constant, or solution step is altered --- mathematical equivalence holds by construction. No judge, no grading model, and no theorem-proving capability is involved. The structural-overlap analysis in \S\ref{sec:dissociation}, which is the paper's headline mechanistic finding, is computed entirely from these deterministic rename maps and the models' own trajectories, with zero LLM annotation at any stage.

\subsection{Robustness of kernel variations}

Every LLM call in the kernel pipeline operates in an information-asymmetric verification role: the judge sees the original problem, the reference solution, the regenerated proof, and the candidate variant, and is asked whether equivalence holds step by step. This is strictly easier than solving the problem from scratch --- the judge need not search the solution space, only check a presented chain. Under an empirical single-judge error rate $\varepsilon \leq 0.04$, the two-in-a-row unanimous rule across $J{=}5$ judges and $T{=}15$ repair iterations yields a soundness bound $\delta \leq 14\,\varepsilon^{KJ} < 10^{-10}$ on a flawed variant surviving the loop. The bound holds at $\varepsilon = 0.08$ with $\delta < 10^{-8}$. The gate is active in practice: $64$ of the $1{,}051$ initial candidates ($6.1\,\%$) failed at least one round and entered repair, and all were eventually accepted (Appendix~\ref{app:repair_stats}). A $10\,\%$ manual audit on accepted variants found no substantive errors.

\subsection{Robustness of grading}

Three independent providers (GPT-4o, Claude-Sonnet-4, Gemini-2.5-flash) grade the same 50 stratified solutions; pairwise Cohen's $\kappa$ ranges from $0.96$ to $1.00$, with agreement on 49--50 of 50 cases. On calculation problems, a strict string-match grader runs in parallel with the LLM grader; disagreement is below $1\,\%$. A separate $10\,\%$ random-sample human audit on the full grading run confirms precision and recall above $97\,\%$.

\subsection{Robustness across datasets}
\label{app:cross_dataset}

We apply the same surface pipeline to 159 MATH Level 4--5 problems \citep{hendrycks:2021} (six models) and to ALG514 \citep{kushman-etal-2014-learning} (one model) to verify that the behavioural asymmetry is not specific to Putnam. All 12 (model, surface-variant) cells on MATH show negative paired drops; 8 of 12 are individually significant by McNemar at $p < 0.05$, with the largest being gpt-4.1-nano on \textsc{GS} ($-21.4$~pp, $p = 1.2 \times 10^{-6}$). On ALG514, accuracy falls from $93.6\%$ to $90.9\%$ under \textsc{DLC} ($p = 0.024$) and to $89.3\%$ under \textsc{GS} ($p = 0.002$; Appendix~\ref{app:alg514}). We claim behavioural transfer of the framework, not mechanism-level reproduction: trajectory-level dissociation on these corpora requires trajectory-capturing reruns and is left to future work.

\subsection{The verification gap as a design principle}

GAP is designed around a structural asymmetry between the solver and the verifier. The models under evaluation must solve a problem from scratch with no reference answer; every LLM inside the pipeline --- generators, judges, graders --- operates with the reference answer, the canonical solution, or both already in hand. Checking a presented proof against a known answer is strictly easier than producing one, so the pipeline does not require frontier-level reasoning from any of its components. This verification gap is what makes multi-judge consensus, bounded repair loops, and cross-provider agreement effective: each mechanism amplifies a base reliability that is already high because the task is verification, not generation.

%% file: sections/appendix/k_additional_evaluations.tex
\section{Additional evaluations and sensitivity analyses}
\label{app:additional}

This appendix tests whether the headline asymmetry --- kernel rewrites
remove far more solvable items than surface renames
(\S\ref{sec:headline}, \S\ref{sec:evidence}) --- survives four
alternative explanations or settings: an increase in intrinsic
difficulty (\S\ref{app:difficulty}), a newer model
(\S\ref{app:newer_model}), stochastic variation across repeated
generations (\S\ref{app:repeated_runs}), and a tool-using agent in
place of a single model call (\S\ref{app:math_agent}). The last three
share the frozen evaluation sample described in
\S\ref{app:frozen_sample}.

\subsection{Is the kernel decrease explained by increased difficulty?}
\label{app:difficulty}

A kernel rewrite changes the setting and the answer, so some rewrites
are plausibly harder than their sources. The paper does not infer its
failure interpretation from the accuracy decrease alone: that
interpretation rests on the trajectory and intervention evidence of
\S\ref{sec:dissociation} and Appendix~\ref{app:trajectory}, and the
kernel--surface gap already holds within every Easy, Medium, and Hard
band (Appendix~\ref{app:loss_by_strata}). The two analyses below
further restrict the comparison to source/kernel pairs on which a
difficulty increase is less plausible.

\paragraph{Outcome-blind difficulty judgement.}
\textsc{DeepSeek-V4-Flash} compared all $1{,}051$ source/kernel pairs.
For each pair it saw both problem statements and both reference
solutions in a randomised A/B order fixed before any accuracy was
computed; it did not see which side was the source, any generation
metadata, or any evaluated-model output. The source appeared on side~A
for $526$ pairs and on side~B for $525$. The judge labelled $537$
pairs as unchanged in intrinsic difficulty, $487$ as kernel-harder,
and $27$ as kernel-easier; the label distribution was not associated
with A/B orientation ($\chi^2$ test, $p=0.357$).

\paragraph{Length-controlled subset.}
Separately, we kept the $333$ pairs for which both the problem-statement
length ratio and the reference-solution length ratio (kernel over
source) lie between $0.5$ and $2$, excluding rewrites whose statement
or solution grew or shrank substantially.

\begin{table}[H]
  \centering
  \small
  \setlength{\tabcolsep}{7pt}
  \caption{Kernel accuracy decrease on two difficulty-controlled
  subsets, macro-averaged over the 18 evaluated models. The last column
  gives the largest mean decrease among the four surface families on
  the same source items.}
  \label{tab:difficulty_controls}
  \newcommand{\hd}[1]{\textbf{\begin{tabular}[b]{@{}c@{}}#1\end{tabular}}}
  \begin{tabular}{@{}lrccc@{}}
    \toprule
    \textbf{Subset} & \textbf{Pairs} &
    \hd{Kernel\\decrease (pp)} &
    \hd{Models with\\a decrease} &
    \hd{Largest surface\\decrease (pp)} \\
    \midrule
    Judged unchanged in difficulty & 537 & 3.43 {\footnotesize$[2.12,\,4.79]$} & 17/18 & 2.03 \\
    Length-controlled              & 333 & 5.59                                 & 17/18 & 3.08 \\
    \bottomrule
  \end{tabular}

  \vspace{0.3em}
  \raggedright
  \footnotesize
  \textit{Note:} bracketed values are a 95\% problem-cluster bootstrap
  interval.
\end{table}

On the $537$ pairs judged unchanged, macro-average accuracy falls from
$21.49\%$ on the source to $18.06\%$ on the kernel rewrite. The
decrease is present for $17$ of $18$ models (exact sign test
$p=1.45\times10^{-4}$), and the four surface-family decreases on the
same items range from $0.17$ to $2.03$\,pp, all smaller than the
kernel decrease. On the $333$ length-controlled pairs, the mean kernel
decrease is $5.59$\,pp, again present for $17$ of $18$ models, with
$11$ of the $18$ per-model exact McNemar tests reaching $p<0.05$; no
surface family exceeds a $3.08$\,pp mean decrease on this subset
(Table~\ref{tab:difficulty_controls}).

Both decreases are smaller than the full-corpus kernel decrease
(median $10.3$\,pp across models; \S\ref{sec:evidence}), which is consistent with
increased difficulty contributing to the \emph{magnitude} of the
aggregate effect. Neither an LLM difficulty judgement nor text length
is a complete psychometric measure of mathematical difficulty, so we
do not claim that the retained pairs are equally difficult. The
narrower conclusion is that the kernel decrease is not confined to
pairs judged harder or to rewrites with unusually long statements or
solutions, and that the kernel--surface ordering persists on both
subsets.

\subsection{Frozen evaluation sample}
\label{app:frozen_sample}

Before any inference for \S\S\ref{app:newer_model}--\ref{app:math_agent},
we froze a sample of $100$ Putnam sources, stratified proportionally by
decade and problem type with a fixed random seed. Each source was
evaluated on the original problem and all five GAP variants with the
zero-shot solver prompt and the proof and calculation grader prompts
of Appendix~\ref{app:prompts}. The grader never saw the transformation
label or the repeat index. API failures are reported as coverage
rather than counted as mathematical errors, and every paired
comparison uses only sources with a valid solution and grade on both
sides.

\subsection{A newer model: DeepSeek-V4-Pro}
\label{app:newer_model}

To verify that the asymmetry is not specific to the 18 main-study
models, we evaluated \textsc{DeepSeek-V4-Pro}, a newer model, on the
frozen 100-source sample (Table~\ref{tab:v4pro}). This is a
verification check on a sample, not a full-corpus evaluation. The kernel decrease
($23.68$\,pp) is $11.92$\,pp larger than the largest surface decrease
(\textsc{GS}, $11.76$\,pp). Among the $76$ valid original/kernel pairs,
$24$ change from original-correct to kernel-wrong, against $6$ in the
reverse direction.

Of the $600$ target calls, $535$ returned a valid solution after the
initial pass and one recovery pass with the same configuration, and
$532$ of these received a valid blind grade. The original problem has
$89/100$ valid solutions; the per-variant counts are listed in the
table.

\begin{table}[H]
  \centering
  \small
  \setlength{\tabcolsep}{4.5pt}
  \caption{\textsc{DeepSeek-V4-Pro} on the frozen 100-source sample.
  Original accuracy is computed on the paired subset for each variant,
  so it differs slightly across rows.}
  \label{tab:v4pro}
  \begin{tabular}{@{}lcccccc@{}}
    \toprule
    \textbf{Variant} & \textbf{Valid solutions} & \textbf{Paired $N$} &
    \textbf{Original (\%)} & \textbf{Variant (\%)} &
    \textbf{Decrease (pp)} & \textbf{Exact McNemar $p$} \\
    \midrule
    \textsc{DL}  & 90/100 & 86 & 59.30 & 55.81 &  3.49 & 0.607 \\
    \textsc{DLC} & 93/100 & 86 & 58.14 & 53.49 &  4.65 & 0.481 \\
    \textsc{DLM} & 94/100 & 87 & 58.62 & 50.57 &  8.05 & 0.189 \\
    \textsc{GS}  & 90/100 & 85 & 58.82 & 47.06 & 11.76 & 0.021 \\
    \textsc{KV}  & 79/100 & 76 & 56.58 & 32.89 & \textbf{23.68} & \textbf{0.0014} \\
    \bottomrule
  \end{tabular}
\end{table}

\subsection{Repeated independent generations}
\label{app:repeated_runs}

The main evaluation queries each model once per item
(Appendix~\ref{app:compute}). To check that the kernel decrease is not
an artefact of a single stochastic generation, we ran
\textsc{DeepSeek-V4-Flash} three times on the frozen sample,
generating every original and kernel answer independently in each run
with unchanged solver and grader prompts. The comparison uses the $87$
sources with valid original and kernel outputs in all three runs
(Table~\ref{tab:repeated_runs}).

\begin{table}[H]
  \centering
  \small
  \caption{Three independent \textsc{DeepSeek-V4-Flash} runs on the
  same $87$ sources.}
  \label{tab:repeated_runs}
  \begin{tabular}{@{}ccccc@{}}
    \toprule
    \textbf{Run} & \textbf{Original (\%)} & \textbf{Kernel (\%)} &
    \textbf{Decrease (pp)} & \textbf{Exact McNemar $p$} \\
    \midrule
    1 & 43.68 & 17.24 & 26.44 & $1.55\times10^{-6}$ \\
    2 & 43.68 & 21.84 & 21.84 & $1.57\times10^{-4}$ \\
    3 & 40.23 & 20.69 & 19.54 & $4.88\times10^{-4}$ \\
    \midrule
    Mean (SD) & & & 22.61 (3.51) & \\
    \bottomrule
  \end{tabular}
\end{table}

Individual answers vary across runs, but kernel accuracy is lower in
every run and every paired difference is significant. This checks one
representative model rather than re-running all 18.

\subsection{Exploratory evaluation of a specialised math agent}
\label{app:math_agent}

GAP studies the robustness of a model under a single invocation with
fixed prompting and grading, including any reasoning or inference-time
computation native to the model. An agent adds an external controller,
multi-call orchestration, iterative search, memory, and tools, so its
behaviour cannot be attributed to the underlying model in the same
way. We therefore report an agent evaluation separately from the main
analysis, as an exploratory application.

We evaluated the open-source Intel DeepMath agent
(\texttt{Intel/deepmath-v1}) with its native prompt, its four official
few-shot traces, its agent loop, and its Python executor, on $40$
stratified sources drawn from the frozen sample. The original problem
and all five variants give $240$ runs, all of which completed and
received condition-blind grades (Table~\ref{tab:math_agent}).

\begin{table}[H]
  \centering
  \small
  \caption{Intel DeepMath agent on $40$ sources from the frozen sample
  (exploratory).}
  \label{tab:math_agent}
  \begin{tabular}{@{}lccc@{}}
    \toprule
    \textbf{Condition} & \textbf{Accuracy (\%)} &
    \textbf{Decrease (pp)} & \textbf{Exact McNemar $p$} \\
    \midrule
    Original     & 27.5 & ---  & ---    \\
    \textsc{DL}  & 17.5 & 10.0 & 0.125  \\
    \textsc{DLC} & 15.0 & 12.5 & 0.0625 \\
    \textsc{DLM} & 22.5 &  5.0 & 0.625  \\
    \textsc{GS}  & 15.0 & 12.5 & 0.0625 \\
    \textsc{KV}  & 20.0 &  7.5 & 0.453  \\
    \bottomrule
  \end{tabular}
\end{table}

Every transformed condition scores below the original, but with $40$
sources no paired difference is significant, and the surface/kernel
ordering differs from the single-model experiments. This is consistent
with the scope distinction above: an external agent loop changes the
evaluated system and may change its failure profile. We do not extend
the paper's claims about LLMs to math agents.

%% file: sections/appendix/c_supplementary.tex
%
%
\section{Supplementary aggregate tables and figures}
\label{app:supplementary}

\subsection{Per-model accuracy with paired significance}
\label{app:aggregate}

Table~\ref{tab:updated_model_scores_pct} reports each model's accuracy
on the original split and on each of the five transformed families,
with paired two-proportion $z$-test significance markers. The table is
the per-model breakdown referenced from
Figure~\ref{fig:accuracy_by_variant} in \S\ref{sec:headline} and
underlies the median-loss and McNemar statistics in
\S\ref{sec:evidence}. Per-cell 95\,\% bootstrap CIs are stored in
\texttt{putnam-bench-anon/bootstrapping/bootstrap\_confidence\_intervals.csv}
(1{,}000 resamples).

\begin{table*}[t]
  \centering
  \small
  \setlength{\tabcolsep}{4.2pt}
  \caption{Model accuracy across families (percent scale). Significance
  markers from paired two-proportion $z$-test against the original
  split. McNemar contrasts on the same pairs are tabulated in
  \texttt{aggregations/mcnemar\_results.csv}.}
  \label{tab:updated_model_scores_pct}
  \newcommand{\pad}{\phantom{$^{***}$}}
  \newcommand{\padone}{$^{*}$\phantom{$^{**}$}}
  \newcommand{\padtwo}{$^{**}$\phantom{$^{*}$}}
  \resizebox{\textwidth}{!}{%
  \begin{tabular}{@{}lcccccc@{}}
    \toprule
    \textbf{Model} &
    \textbf{Original} &
    \textbf{\textsc{DL}} ($\Delta$) &
    \textbf{\textsc{DLC}} ($\Delta$) &
    \textbf{\textsc{DLM}} ($\Delta$) &
    \textbf{\textsc{GS}} ($\Delta$) &
    \textbf{\textsc{KV}} ($\Delta$) \\
    \midrule
    claude-opus-4         & 26.5\pad & 23.0\padone\,(--3.5) & 22.2\padtwo\,(--4.3) & 21.7\padtwo\,(--4.8) & 21.4\padtwo\,(--5.1) & 13.8$^{***}$\,(--12.7) \\
    claude-sonnet-4       & 23.0\pad & 20.6\pad\,(--2.5)    & 19.8\padone\,(--3.2) & 18.6\padtwo\,(--4.4) & 18.1$^{***}$\,(--4.9) & 11.1$^{***}$\,(--11.9) \\
    deepseek-prover       & 15.5\pad & 15.2\pad\,(--0.3)    & 14.0\pad\,(--1.5)    & 12.8\padone\,(--2.7) & 13.7\pad\,(--1.8)     &  9.2$^{***}$\,(--6.3) \\
    gemini-2.5-flash-lite & 19.8\pad & 18.8\pad\,(--0.9)    & 16.1\padtwo\,(--3.7) & 15.8\padtwo\,(--4.0) & 15.1$^{***}$\,(--4.7) &  6.6$^{***}$\,(--13.2) \\
    gemini-2.5-pro        & 78.4\pad & 75.2\padone\,(--3.1) & 74.3\padtwo\,(--4.1) & 72.8$^{***}$\,(--5.6) & 72.9$^{***}$\,(--5.4) & 63.5$^{***}$\,(--14.9) \\
    gemini-2.5-flash      & 42.8\pad & 42.6\pad\,(--0.2)    & 39.0\padone\,(--3.8) & 40.9\pad\,(--1.9)    & 37.6\padtwo\,(--5.2) & 27.6$^{***}$\,(--15.2) \\
    gpt-4.1               & 24.9\pad & 23.4\pad\,(--1.5)    & 21.2\padone\,(--3.7) & 22.0\pad\,(--2.9)    & 22.6\pad\,(--2.3)     & 14.8$^{***}$\,(--10.1) \\
    gpt-4.1-mini          & 28.6\pad & 26.9\pad\,(--1.7)    & 27.1\pad\,(--1.5)    & 24.0\padtwo\,(--4.6) & 25.1\padone\,(--3.4) & 19.2$^{***}$\,(--9.4) \\
    gpt-4.1-nano          &  8.8\pad &  8.9\pad\,(+0.1)     &  6.4\padone\,(--2.4) &  7.3\pad\,(--1.5)    &  6.6\pad\,(--2.2)     &  6.8\pad\,(--2.0) \\
    gpt-4o                &  6.5\pad &  6.3\pad\,(--0.1)    &  5.3\pad\,(--1.1)    &  4.7\padone\,(--1.8) &  4.7\padone\,(--1.8) &  3.0$^{***}$\,(--3.4) \\
    gpt-4o-mini           &  4.3\pad &  3.5\pad\,(--0.8)    &  2.5\padtwo\,(--1.8) &  3.3\pad\,(--1.0)    &  3.2\pad\,(--1.1)     &  1.7$^{***}$\,(--2.6) \\
    grok4                 & 60.1\pad & 59.0\pad\,(--1.1)    & 56.6\pad\,(--3.4)    & 55.9\padone\,(--4.2) & 55.2\padtwo\,(--4.8) & 45.5$^{***}$\,(--14.6) \\
    kimi-k2               & 27.2\pad & 25.8\pad\,(--1.4)    & 23.7\padone\,(--3.5) & 23.8\padone\,(--3.4) & 23.4\padtwo\,(--3.8) & 12.8$^{***}$\,(--14.4) \\
    llama-4               & 15.7\pad & 14.5\pad\,(--1.2)    & 13.0\padone\,(--2.7) & 13.1\padone\,(--2.6) & 13.8\pad\,(--1.9)     &  7.3$^{***}$\,(--8.4) \\
    mistral               &  5.5\pad &  5.5\pad\,(--0.1)    &  5.7\pad\,(+0.1)     &  4.9\pad\,(--0.7)    &  4.2\pad\,(--1.3)     &  3.9\padone\,(--1.6) \\
    o3                    & 51.5\pad & 52.8\pad\,(+1.3)     & 47.6\padone\,(--3.8) & 49.5\pad\,(--2.0)    & 46.8\padtwo\,(--4.7) & 38.6$^{***}$\,(--12.9) \\
    o4-mini               & 41.5\pad & 43.0\pad\,(+1.5)     & 38.3\pad\,(--3.2)    & 38.5\pad\,(--3.0)    & 40.4\pad\,(--1.1)     & 29.1$^{***}$\,(--12.4) \\
    qwen3                 & 28.2\pad & 29.3\pad\,(+1.1)     & 27.3\pad\,(--0.9)    & 26.9\pad\,(--1.4)    & 26.5\pad\,(--1.7)     & 14.9$^{***}$\,(--13.4) \\
    \bottomrule
  \end{tabular}%
  }
  \vspace{0.5em}
  \raggedright
  \textit{Note:} $^{***}p<0.01$, $^{**}p<0.05$, $^{*}p<0.1$.
\end{table*}

\subsection{Capability-loss rate by topic, difficulty, and form}
\label{app:loss_by_strata}

The capability-loss rate $\mathcal{L}=b/(b+s)$ defined in
\S\ref{sec:evidence} is reported here stratified along the three axes
mentioned in \S\ref{sec:headline}: topic
($\mathrm{ALG}$, $\mathrm{ANA}$, $\mathrm{NT}$, $\mathrm{COMB}$,
$\mathrm{GEO}$), difficulty (Easy / Medium / Hard, derived from the
canonical positional index $1$--$2$, $3$--$4$, $5$--$6$), and form
(proof vs.\ calculation). For each stratum we report
$\mathcal{L}^{\mathrm{kern}}$ and $\mathcal{L}^{\mathrm{surf}}$
(the simple mean of the four surface-family rates), as well as each
surface family on its own. Counts are pooled across the 18 evaluated
models: within a stratum we sum $b$ (original solved, variant missed)
and $s$ (both solved) over all (model, problem) pairs and compute
$\mathcal{L}$ on the pooled totals. The per-stratum
kernel\,$-$\,surface gap $\Delta=\mathcal{L}^{\mathrm{kern}}-\mathcal{L}^{\mathrm{surf}}$
is positive in every cell, which is what \S\ref{sec:headline} relies
on when it claims that the gap survives across topic, difficulty, and
form rather than being driven by a single subset. We also include a
``Tail'' difficulty bucket (positional index $7$--$8$) for
completeness; only a handful of years carry items in this range, so
its $b{+}s$ totals are an order of magnitude smaller than the other
buckets and the resulting cell estimates are correspondingly noisier.

\begin{table}[H]
  \centering
  \small
  \setlength{\tabcolsep}{4pt}
  \caption{Capability-loss rate $\mathcal{L}=b/(b+s)$ by topic,
  difficulty, and form, expressed as percentages and pooled across the
  18 evaluated models. $\mathcal{L}^{\mathrm{surf}}$ is the simple
  mean of the four surface-family rates (\textsc{DL}, \textsc{DLC},
  \textsc{DLM}, \textsc{GS}); $\Delta$ is
  $\mathcal{L}^{\mathrm{kern}}-\mathcal{L}^{\mathrm{surf}}$. Computed
  from \texttt{aggregations/all\_results\_summary.csv}.}
  \label{tab:loss_by_strata}
  \begin{tabular}{lcccccccc}
    \toprule
    \textbf{Stratum}
      & \textsc{DL} & \textsc{DLC} & \textsc{DLM} & \textsc{GS}
      & $\mathcal{L}^{\mathrm{surf}}$
      & $\mathcal{L}^{\mathrm{kern}}$
      & $\Delta$ \\
    \midrule
    \multicolumn{8}{l}{\emph{Topic}} \\
    Algebra            & 32.0 & 36.5 & 36.6 & 37.1 & 35.6 & 57.3 & +21.7 \\
    Analysis           & 29.1 & 33.6 & 37.1 & 35.2 & 33.7 & 59.2 & +25.5 \\
    Number Theory      & 31.2 & 35.3 & 33.6 & 37.6 & 34.4 & 57.2 & +22.8 \\
    Combinatorics      & 29.1 & 32.0 & 33.7 & 34.3 & 32.3 & 55.3 & +23.1 \\
    Geometry           & 30.8 & 30.9 & 35.3 & 35.3 & 33.1 & 53.3 & +20.3 \\
    \midrule
    \multicolumn{8}{l}{\emph{Difficulty}} \\
    Easy (idx 1--2)    & 26.3 & 29.9 & 32.3 & 32.2 & 30.2 & 50.5 & +20.3 \\
    Medium (idx 3--4)  & 32.7 & 36.1 & 37.4 & 38.2 & 36.1 & 62.6 & +26.5 \\
    Hard (idx 5--6)    & 33.6 & 36.9 & 38.8 & 38.1 & 36.9 & 59.5 & +22.6 \\
    Tail (idx 7--8)    & 30.0 & 34.0 & 44.7 & 39.3 & 37.0 & 72.0 & +35.0 \\
    \midrule
    \multicolumn{8}{l}{\emph{Form}} \\
    Proof              & 36.2 & 40.3 & 42.0 & 42.8 & 40.3 & 64.8 & +24.5 \\
    Calculation        & 21.3 & 24.0 & 26.9 & 25.6 & 24.4 & 46.5 & +22.1 \\
    \midrule
    All                & 30.0 & 33.6 & 35.7 & 35.7 & 33.8 & 57.3 & +23.5 \\
    \bottomrule
  \end{tabular}
\end{table}

Two patterns are worth noting and are referenced by the main text.
First, the surface\,$\to$\,kernel gap $\Delta$ is positive in every
cell of every axis (range $20.3$--$26.5$\,pp on the principal
buckets), so the headline claim does not depend on any single topic,
difficulty band, or form; Appendix~\ref{app:difficulty} further
restricts the comparison to pairs judged unchanged in difficulty and
to length-controlled pairs. Second, $\mathcal{L}^{\mathrm{surf}}$
itself is monotone within each family in the order
$\textsc{DL}<\textsc{DLC}<\textsc{DLM}\!\approx\!\textsc{GS}$ in
nearly every cell, consistent with the framing in \S\ref{sec:two-families}
that the four surface families are ordered by how aggressively they
break the alignment between identifier and latent role.

\subsection{Supplementary figures}
\label{app:supp_figures}

\begin{figure}[H]
  \centering
  \safeincludegraphics[width=0.65\linewidth]{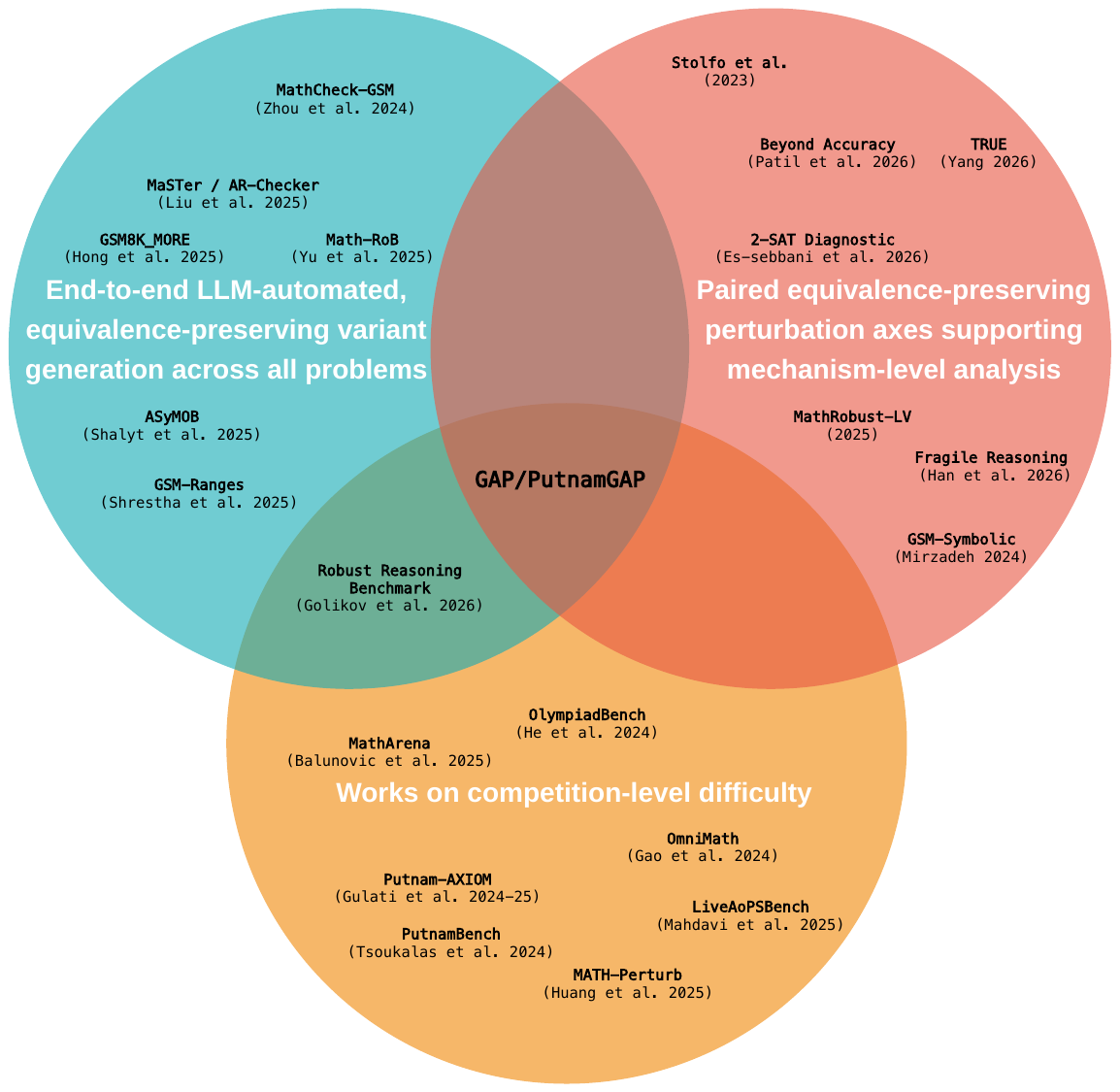}
  \caption{Positioning of GAP at the intersection of three desiderata:
  end-to-end automation with equivalence guarantees, competition-level
  difficulty, and perturbation-axis decomposition for mechanism analysis.
  No prior benchmark occupies all three. Only RRB~\citep{golikov:2026rrb}, which appeared after our work, occupies two of the three.}
  \label{fig:venn}
\end{figure}

\begin{figure}[H]
  \centering
  \safeincludegraphics[width=1\textwidth]{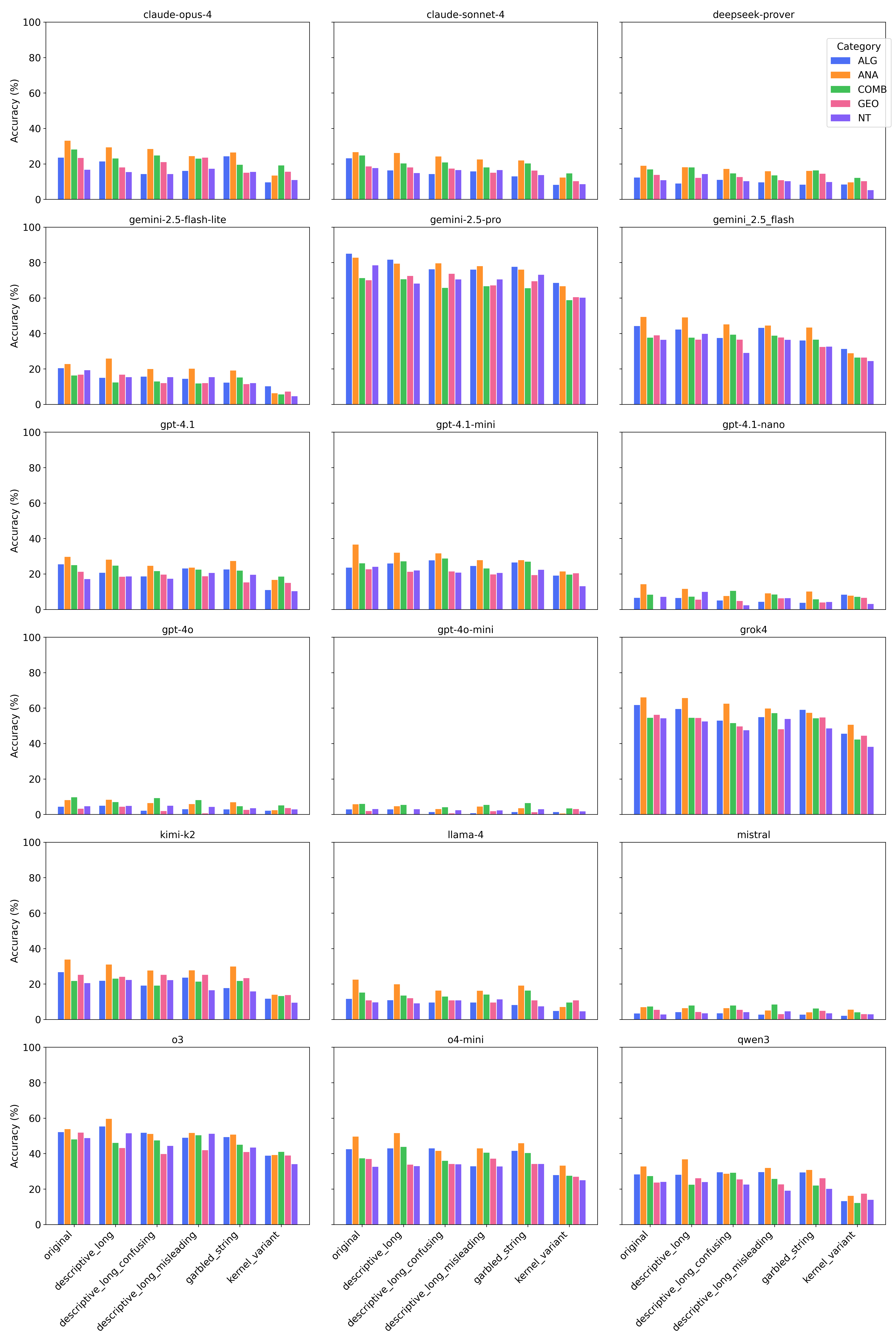}
  \caption{Accuracy on each variant family broken out by topic
  category.}
  \label{fig:accuracy_by_topic}
\end{figure}

\begin{figure}[H]
  \centering
  \safeincludegraphics[width=1\textwidth]{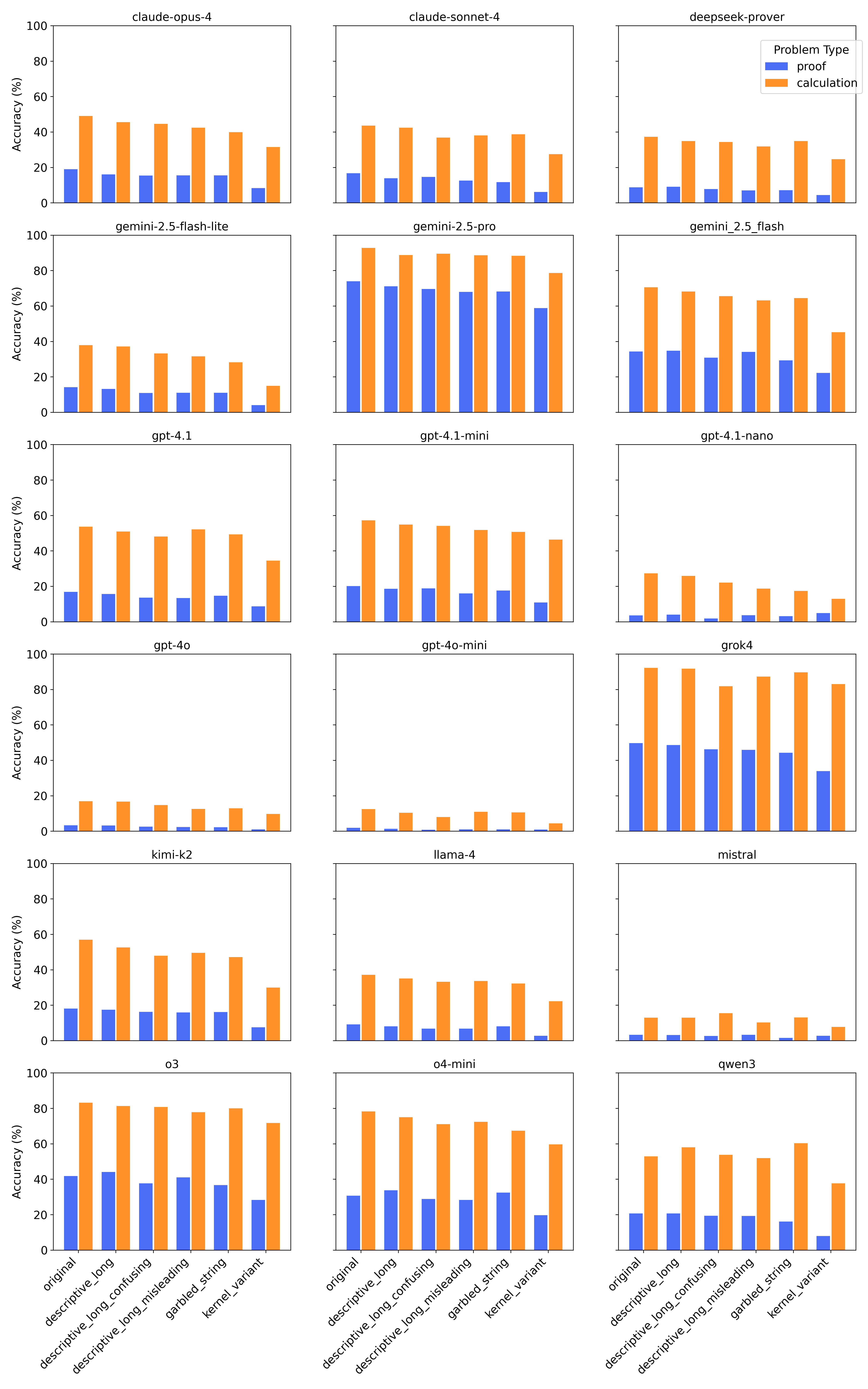}
  \caption{Accuracy on each variant family broken out by problem
  form (proof vs.\ calculation).}
  \label{fig:accuracy_by_form}
\end{figure}

\begin{figure}[H]
  \centering
  \safeincludegraphics[width=0.95\linewidth]{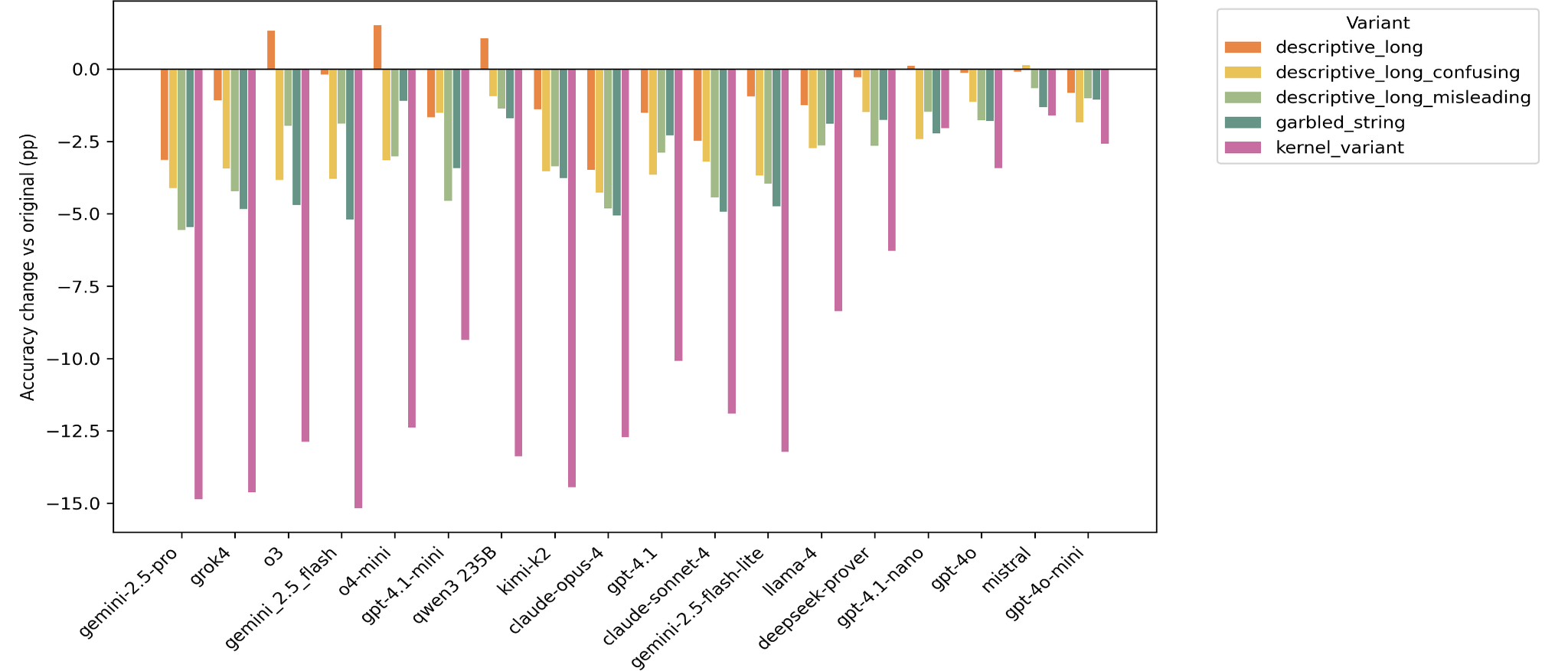}
  \caption{Per-model paired accuracy delta between original and each
  variant family, sorted by aggregate drop. Kernel rewrites occupy
  the rightmost (largest-drop) end for almost every model.}
  \label{fig:variant_diff}
\end{figure}

%% file: sections/appendix/i_trajectory.tex
%
%
%
\section{Trajectory-level mechanism analyses}
\label{app:trajectory}

This appendix develops the trajectory-level analyses summarised in
\S\ref{sec:dissociation} and the rescue contrasts in
\S\ref{sec:evidence}. It contains the canonicalisation and metric
definitions for the paired structural-overlap analysis
(\S\ref{app:overlap}), the rescue protocol with prompt construction,
per-cell tables, paired McNemar contrasts and supporting figures
(\S\ref{app:rescue}), the qualitative archetype audit
(\S\ref{app:rescue_audit}), the collapse failure-mode breakdown
(\S\ref{app:collapse}), the long-form information-asymmetry argument
for the LLM pipeline (\S\ref{app:pipeline_defense}), the
three-provider grader cross-check (\S\ref{app:grader_kappa}),
the ALG514 external-validation summary (\S\ref{app:alg514}), and a
feasibility check of the kernel pipeline outside Putnam
(\S\ref{app:kernel_beyond_putnam}).

\subsection{Paired structural overlap}
\label{app:overlap}

\paragraph{Canonicalisation protocol.}
For each surface variant ($\textsc{DL}$, $\textsc{DLC}$,
$\textsc{DLM}$, $\textsc{GS}$) the dataset stores a deterministic
rename map $\pi:\mathcal{V}_{\text{canonical}}\to\mathcal{V}_{\text{variant}}$
under \texttt{variants[v].map}. Given a model trajectory $t$
(variant-named) and the corresponding original-correct trajectory
$t^{o}$ (canonical-named), we apply word-boundary substitution that
maps each canonical name $v\in\mathcal{V}_{\text{canonical}}$ to a
placeholder \texttt{\_\_V}$_{i}$\texttt{\_\_} in $t^{o}$, and each
variant name $\pi(v)$ to the same placeholder
\texttt{\_\_V}$_{i}$\texttt{\_\_} in $t$. Both trajectories then
live in a shared placeholder space and are directly comparable. The
substitution is applied longest-name-first to avoid prefix
collisions (e.g.\ \texttt{al} eating \texttt{almondtree}).

\paragraph{Drift vs collapse stratification.}
Pairs are stratified into three groups: \emph{stable} (original
correct, variant correct), \emph{brittle-drift} (original correct,
variant wrong, variant trajectory $\geq 200$ characters and at least
$25\%$ of the original-trajectory length), and \emph{brittle-collapse}
(original correct, variant wrong, variant trajectory shorter than
either threshold). Collapse rates are reported separately in
\S\ref{app:collapse}; the structural-overlap analysis uses only
stable and drift pairs.

\paragraph{Metrics.}
We compute six metrics on the canonicalised pair $(t^{o},t)$ and
report Cohen's $d$ for stable vs.\ brittle within each (model,
variant) cell, plus Mann--Whitney $U$ and a 1{,}000-iteration
percentile bootstrap CI on Cohen's $d$. The metrics are: (i) token
Jaccard; (ii) bigram Jaccard; (iii) directional coverage
$|T(t^{o})\cap T(t)|/|T(t^{o})|$; (iv) first-600-token window
Jaccard; (v) first-600-token window bigram Jaccard; (vi) equation-set
Jaccard, where math blocks are extracted via regex (display math,
inline \texttt{\$\dots\$}, \texttt{equation}/\texttt{align}
environments) and trivial fragments shorter than 8 characters are
filtered. Metrics (i) and (ii) are the headline; the remainder serve
as robustness checks.

\paragraph{Random-pairing noise floor.}
For each (model, variant) cell we additionally compute the metric on
$N=400$ random pairs $(t^{o}_{i}, t_{j})$ drawn from the stable group
with $i\neq j$. This shuffles the same-problem signal and provides a
baseline that the in-pair Jaccard must exceed for the metric to be
meaningful. Across all 72 surface cells, the random-pairing median
Jaccard is approximately $0.18$, versus a same-problem stable median
in the range $0.36$--$0.59$ and a brittle median in the range
$0.32$--$0.53$. The same-problem signal is therefore roughly twice
the noise floor, and the stable--brittle gap is meaningful within
that signal envelope.

\paragraph{Reference-frame dissociation (full table).}
Table~\ref{tab:dissociation_full} reports the headline metric (token
Jaccard) under three reference frames: the model's own
original-correct trajectory after rename canonicalisation
(\emph{self-anchor}, surface only), the dataset's canonical variant
solution (\emph{canonical-anchor}, both surface and \textsc{KV}),
and the random-pairing noise floor.

\begin{table}[H]
  \centering
  \small
  \caption{Stable-vs-brittle Cohen's $d$ for token Jaccard under
  three reference frames, averaged across cells. Surface = mean over
  18 models $\times$ 4 variants (72 cells); \textsc{KV} = mean over
  18 models. The self-anchor effect ($d=+0.75$) collapses under
  canonical anchor ($d=+0.23$) on surface, demonstrating that
  brittle surface failures are characterised by drift from the
  model's own representation, not from the canonical right answer.
  Kernel failures show the opposite pattern.}
  \label{tab:dissociation_full}
  \begin{tabular}{lcc}
    \toprule
    \textbf{Reference frame} & \textbf{Surface} & \textbf{\textsc{KV}} \\
    \midrule
    Model's own original (renamed)              & $\mathbf{+0.75}$ & --- \\
    Dataset canonical variant solution          & $+0.23$          & $\mathbf{+0.35}$ \\
    Random-pairing noise floor                  & $\approx 0$      & $\approx 0$ \\
    \midrule
    Cells direction-positive (self-anchor)      & $72/72$          & --- \\
    Cells $p<0.05$ (Mann--Whitney, self-anchor) & $69/72$          & --- \\
    Cells $p<10^{-6}$ (self-anchor)             & $40/72$          & --- \\
    \bottomrule
  \end{tabular}
\end{table}

\paragraph{Variant-length tertile stratification.}
A natural concern is that the surface self-anchor effect could be
inflated by length: if brittle trajectories are systematically
shorter, an absolute Jaccard gap may be partly a length artefact. To
remove this, we stratify each cell into variant-length tertiles
(short, medium, long) and recompute Cohen's $d$ within each tertile.
The shortest tertile retains $72/72$ direction-positive cells with
mean $d = +0.84$; the medium and long tertiles are also positive in
every cell, with means $d=+0.71$ and $d=+0.69$ respectively. The
self-anchor signal is therefore not a length artefact: the shortest
tertile, which is the worst case for length confound, carries the
strongest separation.

\paragraph{Per-cell heatmap.}
The cell-level Cohen's $d$ values (self-anchor surface and
canonical-anchor \textsc{KV}) are visualised by
\texttt{figures/fig\_heatmap\_dissociation.pdf} (used in
\S\ref{sec:dissociation}). Models are sorted by mean surface effect
size; among the 18 models, \textsc{gpt-4.1-mini} has the largest
surface effect ($d\approx 0.91$) and \textsc{grok4} has the smallest
($d\approx 0.57$). Effect size is broadly inversely correlated with
raw model strength, consistent with the interpretation that stronger
models retain more of their internal representation under renaming.

\subsection{Repairability under prefix restoration}
\label{app:rescue}

\paragraph{Model selection.}
We rerun the rescue experiment on four representative models chosen
to span provider families and capability tiers: \textsc{gpt-4.1-mini}
(OpenAI, mid-tier, drift $d=+0.91$), \textsc{gpt-4o-mini} (OpenAI,
smaller, drift $d=+0.87$), \textsc{claude-sonnet-4} (Anthropic,
frontier, drift $d=+0.73$), and \textsc{gemini-2.5-flash} (Google,
smaller, drift $d=+0.63$). The grader is \textsc{gpt-4o} for all
conditions, matching the original solver/grader pipeline.

\paragraph{T2 truncation rule.}
\label{app:rescue_t2_rule}
For each prefix source (the model's own original-correct trajectory
or the dataset's canonical variant solution) we apply an
event-boundary truncation labelled \texttt{T2}: keep characters up
to and including the first display-math block (regex
\texttt{\$\$.+?\$\$}, \texttt{\textbackslash[\dots\textbackslash]},
or \texttt{\textbackslash begin\{equation\}\dots}); failing that, up
to the first inline relation containing \texttt{=}, \texttt{<},
\texttt{>}, \texttt{\textbackslash le}, \texttt{\textbackslash ge},
or \texttt{\textbackslash equiv} (excluding mere variable
definitions of the form ``let $x=\dots$''). A leakage filter then
further truncates the prefix at the earliest occurrence of any of:
\texttt{\textbackslash boxed}, ``therefore'', ``hence'', ``we
conclude'', ``the answer'', ``we obtain'', ``thus'', ``it
suffices'', ``we have proved'', ``we have shown'', or the dataset's
recorded \texttt{final\_answer} string (when its length is between
$8$ and $200$ characters). The result is capped at $2{,}400$
characters and required to be at least $50$ characters; otherwise
the case is discarded. For the \texttt{own\_T2} arm we additionally
apply the dataset's rename map to translate canonical variable
names into the variant naming. The resulting prefix is injected via
the prompt template in Appendix~\ref{app:rescue_prompt}.

\paragraph{Per-cell rebound rates.}
Table~\ref{tab:rescue_per_cell} reports rebound rates per (model,
variant, condition) triple with raw counts. Sample sizes are 30 per
cell where available; \textsc{gpt-4o-mini} has fewer flip cases for
some surface variants because its baseline accuracy is low and thus
fewer original-correct items are available to flip.

\begin{table}[H]
  \centering
  \scriptsize
  \setlength{\tabcolsep}{4pt}
  \caption{Per-cell rebound rates ($k/n$) under three prefix
  conditions across 4 representative models and 5 variants.
  \textsc{KV} has only two conditions because the model's own
  original trajectory does not apply (the kernel variant changes the
  mathematics).}
  \label{tab:rescue_per_cell}
  \begin{tabular}{llccc}
    \toprule
    \textbf{Model} & \textbf{Variant} & \textbf{null} & \textbf{canonical\_T2} & \textbf{own\_T2} \\
    \midrule
    \multirow{5}{*}{claude-sonnet-4}
       & \textsc{DL}  & 10/30 (33\%) & 16/28 (57\%) &  9/30 (30\%) \\
       & \textsc{DLC} &  7/30 (23\%) & 13/28 (46\%) & 11/30 (37\%) \\
       & \textsc{DLM} &  8/30 (27\%) & 12/24 (50\%) &  7/30 (23\%) \\
       & \textsc{GS}  &  8/30 (27\%) &  8/28 (29\%) &  5/30 (17\%) \\
       & \textsc{KV}  &  3/30 (10\%) &  4/29 (14\%) & --- \\
    \midrule
    \multirow{5}{*}{gemini-2.5-flash}
       & \textsc{DL}  &  8/30 (27\%) & 14/30 (47\%) &  9/30 (30\%) \\
       & \textsc{DLC} &  8/30 (27\%) &  9/26 (35\%) &  7/30 (23\%) \\
       & \textsc{DLM} & 10/30 (33\%) & 10/28 (36\%) &  8/30 (27\%) \\
       & \textsc{GS}  &  7/30 (23\%) & 12/27 (44\%) &  8/30 (27\%) \\
       & \textsc{KV}  &  5/30 (17\%) &  6/30 (20\%) & --- \\
    \midrule
    \multirow{5}{*}{gpt-4.1-mini}
       & \textsc{DL}  &  8/30 (27\%) & 10/29 (35\%) &  9/30 (30\%) \\
       & \textsc{DLC} &  5/30 (17\%) & 12/28 (43\%) &  4/30 (13\%) \\
       & \textsc{DLM} &  6/30 (20\%) &  8/28 (29\%) &  8/30 (27\%) \\
       & \textsc{GS}  &  7/30 (23\%) & 14/29 (48\%) &  8/30 (27\%) \\
       & \textsc{KV}  &  3/30 (10\%) &  8/28 (29\%) & --- \\
    \midrule
    \multirow{5}{*}{gpt-4o-mini}
       & \textsc{DL}  &  2/15 (13\%) &  5/15 (33\%) &  6/15 (40\%) \\
       & \textsc{DLC} &  6/24 (25\%) &  9/22 (41\%) & 13/24 (54\%) \\
       & \textsc{DLM} &  4/23 (17\%) &  8/21 (38\%) & 11/23 (48\%) \\
       & \textsc{GS}  &  4/20 (20\%) &  2/18 (11\%) &  4/20 (20\%) \\
       & \textsc{KV}  &  8/30 (27\%) &  9/29 (31\%) & --- \\
    \bottomrule
  \end{tabular}
\end{table}

\paragraph{Pooled rates and Wilson CIs.}
Table~\ref{tab:rescue_pooled} reports the pooled-across-models
rebound rates and 95\% Wilson confidence intervals.

\begin{table}[H]
  \centering
  \small
  \caption{Pooled rebound rates with 95\% Wilson confidence
  intervals across the 4 rescue models. Surface variants gain
  $+12$ to $+19$\,pp from \texttt{canonical\_T2}; \textsc{KV} gains
  only $+7.4$\,pp.}
  \label{tab:rescue_pooled}
  \begin{tabular}{lccc}
    \toprule
    \textbf{Variant} & \textbf{null} & \textbf{canonical\_T2} & \textbf{own\_T2} \\
    \midrule
    \textsc{DL}  & 26.7 [19.1, 35.8] & \textbf{44.1 [34.9, 53.8]} & 31.4 [23.3, 40.8] \\
    \textsc{DLC} & 22.8 [16.1, 31.3] & \textbf{41.3 [32.4, 51.0]} & 30.7 [23.0, 39.7] \\
    \textsc{DLM} & 24.8 [17.7, 33.5] & \textbf{37.6 [28.8, 47.4]} & 30.1 [22.4, 39.1] \\
    \textsc{GS}  & 23.6 [16.7, 32.4] & \textbf{35.3 [26.7, 44.9]} & 22.7 [15.9, 31.4] \\
    \textsc{KV}  & 15.8 [10.4, 23.4] & \textbf{23.3 [16.5, 31.7]} & --- \\
    \bottomrule
  \end{tabular}
\end{table}

\paragraph{Pooled paired McNemar contrasts.}
Table~\ref{tab:rescue_mcnemar} reports the three pairwise McNemar
contrasts pooled across all surface cells (and separately for
\textsc{KV}). $b$ counts items where condition A is correct and
condition B is incorrect; $c$ counts the reverse.

\begin{table}[H]
  \centering
  \small
  \caption{Pooled paired McNemar contrasts across all 4 surface
  variants $\times$ 4 models. \texttt{canonical\_T2} significantly
  outperforms both \texttt{null} and \texttt{own\_T2};
  \texttt{own\_T2} does not significantly outperform \texttt{null}.
  The \textsc{KV} contrast is much weaker.}
  \label{tab:rescue_mcnemar}
  \begin{tabular}{lccccc}
    \toprule
    \textbf{Contrast (A $>$ B)} & $b$ & $c$ & \textbf{OR} & \textbf{McNemar }$p$ & $n_{\text{disc}}$ \\
    \midrule
    canonical\_T2 $>$ null              & 90 & 28 & $\mathbf{3.21}$ & $\mathbf{8.9{\times}10^{-9}}$ & 118 \\
    own\_T2 $>$ null                    & 63 & 44 & 1.43            & $8.1{\times}10^{-2}$         & 107 \\
    canonical\_T2 $>$ own\_T2           & 75 & 36 & 2.08            & $\mathbf{2.7{\times}10^{-4}}$ & 111 \\
    \midrule
    \textsc{KV}: canonical\_T2 $>$ null & 15 &  6 & 2.50            & $7.8{\times}10^{-2}$          & 21  \\
    \bottomrule
  \end{tabular}
\end{table}

\paragraph{Prefix-length control.}
The canonical\_T2 $>$ own\_T2 contrast could in principle reflect
nothing more than the expert prefix carrying more text. Both prefixes
are item-specific partial solutions in the variant's vocabulary and
are cut by the same T2 rule, so neither reveals the final answer.
Across the $409$ paired surface cases, canonical\_T2 recovers $162$
cases and own\_T2 recovers $123$, a difference of $9.54$\,pp (exact
McNemar $p=2.7\times10^{-4}$), even though canonical prefixes are
\emph{shorter} on average ($491$ versus $546$ characters). Restricting
to the $87$ cases whose two prefix lengths differ by at most $25\%$,
canonical\_T2 recovers $42/87$ ($48.3\%$) and own\_T2 recovers $29/87$
($33.3\%$), a difference of $14.94$\,pp (exact McNemar $p=0.0072$).
Prefix length therefore does not explain the contrast. What remains
is a difference in content --- the expert prefix supplies the opening
of the reference proof plan --- which is precisely what the
intervention is designed to vary and cannot be equalised without
removing it. We accordingly read the contrast as evidence that the
two failure families are recoverable to different degrees, not as
identification of a unique internal mechanism.

\paragraph{Supporting figures.}

\begin{figure}[H]
  \centering
  \safeincludegraphics[width=\linewidth]{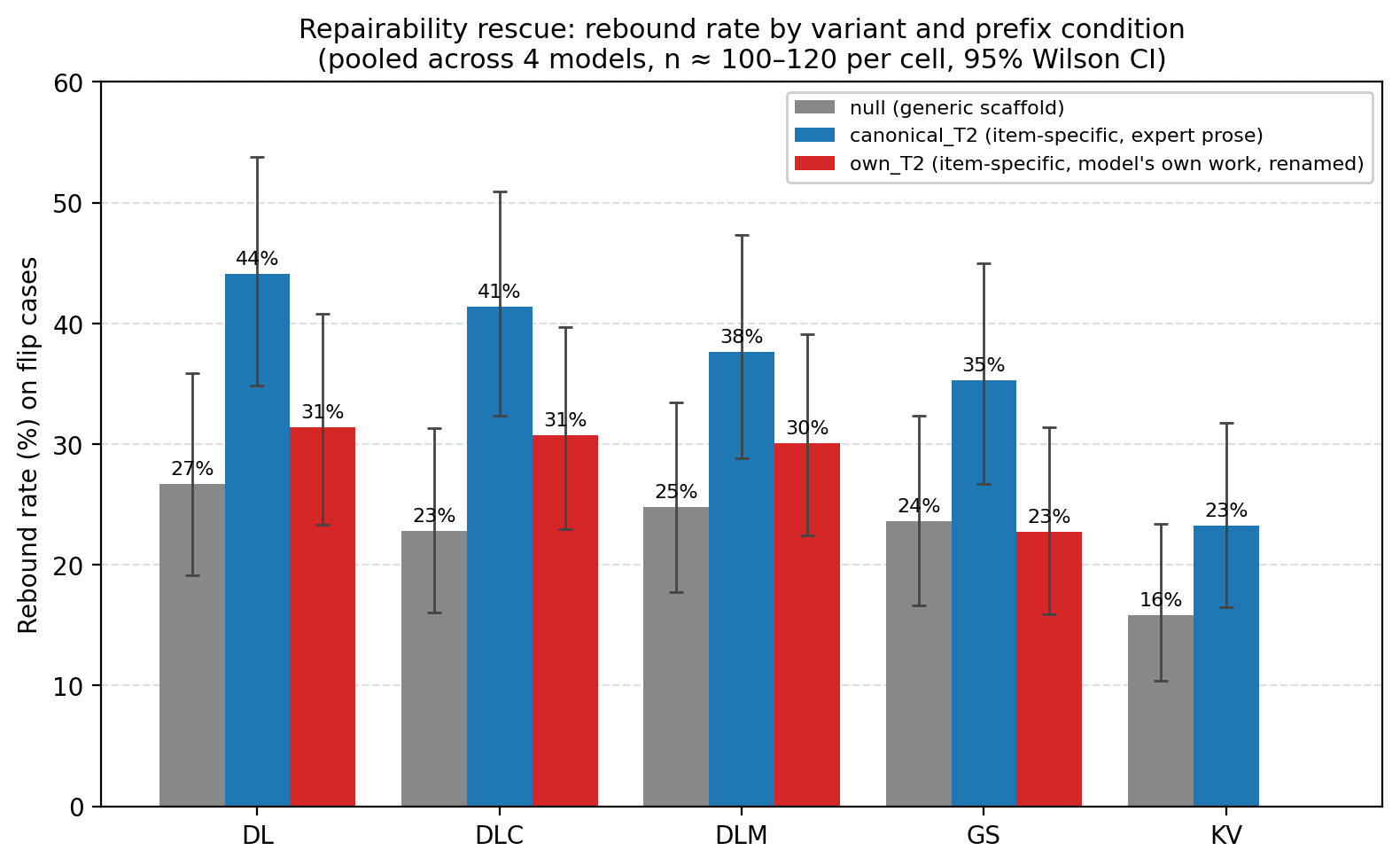}
  \caption{Rebound rate by variant family and prefix condition,
  pooled across the 4 rescue models ($n\!\approx\!100$--$120$ per
  cell, 95\% Wilson CI). \texttt{canonical\_T2} dominates on every
  surface variant; the kernel-variant rebound is consistently
  smaller than the surface rebound.}
  \label{fig:rescue_rebound}
\end{figure}

\begin{figure}[H]
  \centering
  \safeincludegraphics[width=\linewidth]{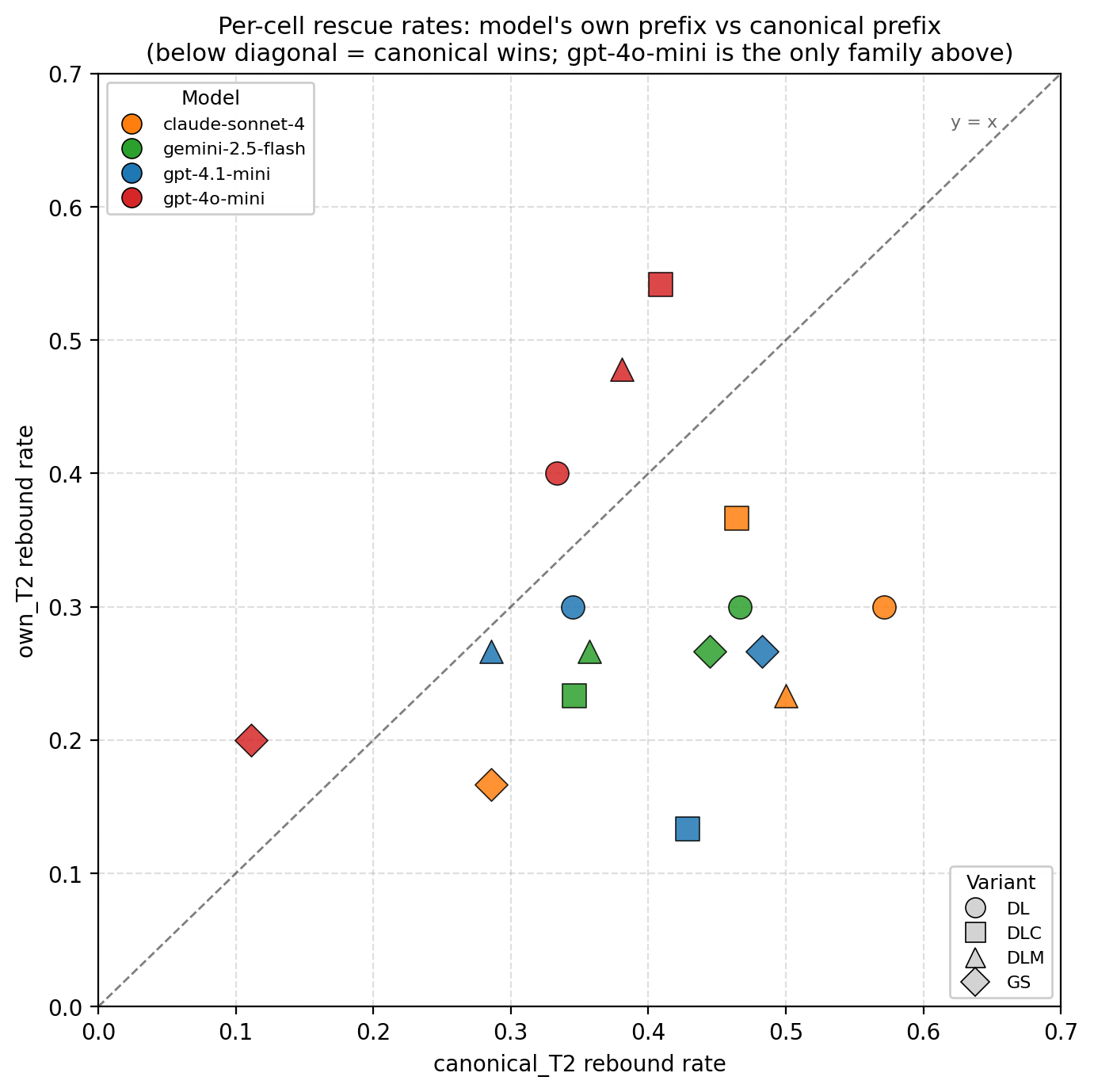}
  \caption{Per-cell rescue rates: \texttt{own\_T2} vs
  \texttt{canonical\_T2} for each (model, surface variant). Most
  cells lie below the diagonal (canonical wins). The four
  \textsc{gpt-4o-mini} cells are the only points above the diagonal,
  indicating that for the weakest model the model's own renamed
  prefix outperforms expert canonical prose.}
  \label{fig:own_vs_canonical}
\end{figure}

\paragraph{Style-matched rescue: the \textsc{gpt-4o-mini} exception.}
For \textsc{gpt-4o-mini}, the only weak model in the rescue panel,
\texttt{own\_T2} outperforms \texttt{canonical\_T2} on three of four
surface variants (\textsc{DL}: 40\% vs 33\%; \textsc{DLC}: 54\% vs
41\%; \textsc{DLM}: 48\% vs 38\%; \textsc{GS}: 20\% vs 11\%). The
pattern is consistent and inverts the result for the three stronger
models in the panel. We read it as evidence that the rescue
mechanism is not merely ``any correct setup helps''; the helpful
object is a setup whose \emph{prose style} the model can extend.
Stronger models can absorb expert canonical prose and continue from
it, while \textsc{gpt-4o-mini} apparently cannot match the
canonical's density and benefits more from its own (simpler) prose.
This motivates the plan-first decoding direction in
Appendix~\ref{app:recommendations}.

\subsection{Qualitative archetypes of canonical-prefix rescues}
\label{app:rescue_audit}

A purely quantitative rebound delta does not say \emph{what}
canonical prefixes supply that own prefixes lack. To approach this
we manually audited 12 paired
\texttt{(canonical\_T2, own\_T2)} cases drawn from the surface arm,
stratified across the four rescue models and four variant families.
Five recurring archetypes emerged; each archetype names what the
canonical prefix introduces beyond the model's own renamed prefix.

\begin{enumerate}[nosep,leftmargin=1.4em]
  \item \textbf{Closed-form ansatz.} The canonical prefix names the
        functional family of the answer (e.g.\ ``the answer is of
        the form $a+b\sqrt{n}$''), reframing the task from discovery
        to verification of constants.
  \item \textbf{Structural commitment.} The canonical prefix
        commits to a global structural choice (e.g.\ ``parametrise
        the curve by $x=t^{2}$, $y=t^{3}$''; ``apply Lagrange
        multipliers'') that the model's own original prefix had not
        yet declared by the \texttt{T2} cut.
  \item \textbf{Ordering of decomposition.} The canonical prefix
        fixes the order in which sub-problems are addressed (e.g.\
        ``handle the symmetric case first; the general case follows
        by reduction''); the own-prefix tends to start with the
        general case and stall.
  \item \textbf{Choice of working basis.} The canonical prefix
        picks an algebraic basis or coordinate system (e.g.\
        polynomial vs.\ matrix form; Cartesian vs.\ polar) that the
        own-prefix had not yet committed to.
  \item \textbf{Dimensional or sign discipline.} The canonical
        prefix flags an invariant the model needs to respect (e.g.\
        a sign condition on a discriminant, a dimensional check on
        a guard inequality); the own-prefix omits the flag.
\end{enumerate}

In the audited 12 cases, every successful canonical-rescue case
fits at least one archetype, and 7 of the 12 fit two simultaneously
(``ansatz $+$ structural commitment'' is the most common pair).
$55$--$60\%$ of surface flips remain unrescued under
\texttt{canonical\_T2} even when one of the five archetypes is
present, so the canonical scaffold is necessary but not sufficient.

\subsection{Collapse failure mode}
\label{app:collapse}

\paragraph{Definition.}
A flip case is classified as \emph{collapse} if the variant
trajectory contains fewer than 200 characters or fewer than $25\%$
of the characters of the model's original-correct trajectory for the
same problem. The remaining flip cases are classified as
\emph{drift}. Collapse and drift are analysed separately because
they reflect distinct failure regimes (the model gives up vs.\ the
model continues but misroutes).

\paragraph{Per-cell collapse rates.}
Table~\ref{tab:collapse} reports the collapse rate (collapse / total
brittle) for each (model, surface variant) cell across all 18
models. Two patterns are visible: (a) \textsc{claude-opus-4} reaches
$63\%$ on \textsc{DLM}, suggesting that this model handles confusing
surface renaming by abandoning the attempt rather than misrouting;
(b) collapse rate is highly variant-dependent within a model, with
\textsc{DLC} and \textsc{DLM} producing the most collapses,
consistent with the intuition that semantically misleading or
confusing names trigger refusal-style behaviour.

\begin{table}[H]
  \centering
  \scriptsize
  \setlength{\tabcolsep}{4pt}
  \caption{Collapse rate (collapse / total brittle, percentage) per
  (model, variant) cell. ``---'' indicates a cell with insufficient
  flip cases.}
  \label{tab:collapse}
  \begin{tabular}{lcccc}
    \toprule
    \textbf{Model} & \textbf{\textsc{DL}} & \textbf{\textsc{DLC}} & \textbf{\textsc{DLM}} & \textbf{\textsc{GS}} \\
    \midrule
    claude-opus-4         &  2 &  1 & \textbf{63} & \textbf{26} \\
    claude-sonnet-4       &  1 &  1 &  4 &  3 \\
    deepseek-prover       &  0 &  2 &  0 &  2 \\
    gemini-2.5-flash-lite &  5 & 10 & 13 &  7 \\
    gemini-2.5-pro        &  5 &  8 & 10 &  6 \\
    gemini-2.5-flash      &  4 &  6 &  4 &  6 \\
    gpt-4.1               &  0 &  1 &  0 &  1 \\
    gpt-4.1-mini          &  1 &  1 &  0 &  0 \\
    gpt-4.1-nano          & 11 &  3 & 19 & 17 \\
    gpt-4o                &  0 &  0 &  0 &  0 \\
    gpt-4o-mini           &  0 &  0 &  0 &  0 \\
    grok4                 &  4 & \textbf{35} &  3 &  3 \\
    kimi-k2               &  2 & 12 &  6 & 11 \\
    llama-4               &  2 &  0 &  0 &  0 \\
    mistral               &  0 &  0 &  6 &  3 \\
    o3                    &  1 &  6 &  5 &  3 \\
    o4-mini               &  0 &  0 &  0 &  0 \\
    qwen3                 &  5 & \textbf{26} & \textbf{19} & \textbf{16} \\
    \bottomrule
  \end{tabular}
\end{table}

\paragraph{Why collapse is reported separately.}
Lumping collapse and drift together produces misleading aggregate
statistics: a model with a $63\%$ collapse rate on a variant family
is solving a fundamentally different problem from a model with a
$5\%$ collapse rate but a heavy drift signal. The two regimes also
call for different mitigations: collapse responds to refusal-aware
fine-tuning and prompting, while drift responds to
representation-stability and plan-stability interventions
(Appendix~\ref{app:recommendations}).

\subsection{Information-asymmetry defence for the LLM pipeline}
\label{app:pipeline_defense}

\paragraph{The structural analysis is zero-LLM by construction.}
The headline structural-overlap finding -- $d=+0.75$ self-anchor vs
$d=+0.23$ canonical-anchor across $72/72$ surface cells -- is
computed entirely without LLM grading or labelling. The rename maps
are read directly from the dataset's deterministic \texttt{map}
field; the metric is token Jaccard on canonicalised text; the
stratification into stable and brittle pairs uses the original
solver/grader correctness label that the paper already requires for
any analysis. No additional LLM is invoked. The main mechanistic
conclusion of the paper is therefore independent of LLM
grader/generator quality.

\paragraph{Solver vs verifier task asymmetry.}
Where GAP does invoke LLMs (kernel-variant generation,
kernel-variant verification, and grading), they operate in an
information-asymmetric verification role rather than a solver role.
\begin{itemize}[leftmargin=*,nosep]
  \item The \emph{solver} task receives only the problem statement
        $Q$ and must produce the answer $A$ from scratch.
  \item The \emph{verifier} task receives $Q$, the reference answer
        $A_{\text{ref}}$, and a candidate variant or candidate
        solution; it must judge equivalence or correctness.
\end{itemize}
Verification under ground truth is fundamentally easier than
from-scratch generation because the verifier can compare against
$A_{\text{ref}}$ and need not search the solution space. An LLM
that is not strong enough to solve a Putnam problem can still
reliably check whether two formulations are equivalent given the
solution. This is why our pipeline can use mid-tier LLMs as judges
and still produce reliable variants, and why the capability tier of
any single judge is not the limiting factor.

\paragraph{Multi-judge protocol.}
For kernel-variant generation we use a 5-judge unanimous protocol
with explicit repair loops (Algorithm~\ref{alg:loop}): a candidate
variant is accepted only if all five judges agree it is
mathematically equivalent to the original on a battery of probe
inputs; otherwise the proposer revises and the cycle repeats up to
$T=15$ iterations. Reliability of GAP comes from the protocol
architecture -- multi-judge consensus, repair loops, cross-provider
agreement, spot audits, and zero-LLM structural metrics for the
surface-side headline -- rather than from the capability tier of
any single model in the pipeline.

\subsection{Three-provider grader cross-check}
\label{app:grader_kappa}

\begin{table}[H]
  \centering
  \small
  \caption{Three-provider grader cross-check Cohen's $\kappa$ on 50
  stratified mixed solutions. All three providers agree on
  $49$--$50$ of $50$ cases.}
  \label{tab:grader_kappa}
  \begin{tabular}{lc}
    \toprule
    \textbf{Provider pair} & \textbf{Cohen's }$\kappa$ \\
    \midrule
    GPT-4o vs Claude-Sonnet-4    & 1.00 \\
    GPT-4o vs Gemini-2.5-flash   & 0.96 \\
    Claude-Sonnet-4 vs Gemini-2.5-flash & 0.96 \\
    \bottomrule
  \end{tabular}
\end{table}

The high $\kappa$ values across providers indicate that the grading
signal is not driven by any one model's idiosyncratic style. This
is the reference cited from \S\ref{sec:reliability} and
\S\ref{sec:putnamgap}.

\subsection{External validation on ALG514}
\label{app:alg514}

To check that GAP's surface-renaming protocols transfer beyond
Putnam-style competition mathematics, we applied the \textsc{DLC}
and \textsc{GS} renamers to ALG514~\citep{kushman-etal-2014-learning},
an algebra word-problem dataset with a different format. Accuracy
decreased from a base of $93.6\%$ to $\textsc{DLC}\;90.9\%$
($\Delta=-2.7$\,pp) and $\textsc{GS}\;89.3\%$
($\Delta=-4.3$\,pp); McNemar tests give $b=24$, $c=10$, $p=0.024$
for base vs.\ \textsc{DLC} and $b=35$, $c=13$, $p=0.002$ for base
vs.\ \textsc{GS}. The drops are statistically significant. The
behavioural transfer is the claim cited in \S\ref{sec:discussion};
trajectory-level dissociation on ALG514 would require additional
trajectory-capturing reruns and is left to future work.

\subsection{Kernel-pipeline feasibility beyond Putnam}
\label{app:kernel_beyond_putnam}

ALG514 tests only the surface side. As a feasibility check for the
kernel side, we ran the unchanged five-stage kernel prompts and the
same five-judge, two-in-a-row protocol on three preselected short
problems from outside Putnam: a Level-5 MATH algebra problem, a
Level-5 MATH number-theory problem~\citep{hendrycks:2021}, and a
GSM8K arithmetic problem~\citep{Cobbe:2021}, using
\textsc{Gemini-2.5-Flash} as the backend. All three runs produced
mathematically correct, non-identical variants that kept the source
solution method and passed two consecutive $5/5$ rounds without
repair. For example, the MATH congruence $617n\equiv943n\pmod{18}$ was
rewritten as $73n\equiv293n\pmod{24}$: the answer changes from $9$ to
$6$, while the solution keeps the same subtract--reduce--coprimality
argument.

Three items demonstrate that the kernel pipeline can operate outside
the Putnam archive; they are not a population-level generalisation
result. The trajectory-level surface/kernel dissociation remains
established on PutnamGAP only.

\subsection{Pilot observation: spontaneous variant-to-canonical normalisation}
\label{app:normalization}

During development of the rescue pipeline we observed an
instructive phenomenon on a subset of \texttt{own\_T2} cases. When
the rename map for an item assigns variant names that overlap with
English vocabulary -- for example a parameter \texttt{a} renamed to
\texttt{coefalpha} on item 1989-B-1, where the prose contains the
article ``a'' in ``a randomly chosen point'' -- the mechanical
word-boundary substitution rewrites the article as well, producing
prose like ``coefalpha randomly chosen point P = (horizcoor,
verticcoor)''. The model often silently restores the article to
``a'' and replaces the variant variable names with their canonical
originals during continuation, indicating an implicit attempt to
repair semantically broken prose. Across 80 \texttt{own\_T2}
trajectories the median variant-name preservation ratio is $1.0$,
so this behaviour is the exception rather than the rule, but the
existence of the behaviour is itself diagnostic: when the prefix is
internally inconsistent, the model's recovery move is to
canonicalise, which is consistent with the broader claim that
surface failures involve fragile bindings between renamed names and
the latent variable identities.

%% file: sections/appendix/a_pipelines.tex
%
%
\section{Variant-generation pipelines and soundness bound}
\label{app:pipelines}

This appendix gives the long-form description of the surface and
kernel pipelines summarised in \S\ref{sec:pipelines}, together with
the soundness analysis backing the $\delta < 10^{-10}$ claim used
there.

\subsection{Surface pipeline}
\label{app:surf_pipeline}

The surface pipeline applies a bijective rename $\tau\in\mathcal{T}^{\mathrm{surf}}$
to the free-variable and fixed-parameter tokens of an item. Scientific
constants such as $\pi$ and $e$ are excluded from the rename
inventory. A single \textsc{o3} call proposes a candidate replacement
conditioned on the token's role (``free variable'' vs.\ ``fixed
parameter''); a deterministic post-validator rejects any name that
collides with another identifier already present in the problem, and
the accepted candidate is routed by a lightweight tagger to one of
four family bins. We instantiate one variant per item per family,
yielding $1{,}051\times 4 = 4{,}204$ surface items.

The four families span a spectrum of lexical informativeness:

\begin{enumerate}[nosep,leftmargin=1.4em]
  \item \textbf{Descriptive-Long (\textsc{DL}).} A single, semantically
        helpful English phrase (e.g.\ \texttt{populationDensity}). \textsc{DL}
        is the maximally-informative end of the spectrum.
  \item \textbf{Descriptive-Long-Confusing (\textsc{DLC}).} A
        concatenation of $2$--$5$ unrelated common words
        (e.g.\ \texttt{walnutVioletTerrace}). The composition is
        grammatical English but supplies no role hint.
  \item \textbf{Descriptive-Long-Misleading (\textsc{DLM}).} A phrase
        built from mathematical jargon that suggests a different
        concept than the one the variable plays
        (e.g.\ \texttt{primeFieldOrder} used as a real variable),
        designed to probe whether models latch onto spurious lexical
        cues.
  \item \textbf{Garbled-String (\textsc{GS}).} A 4--16 character
        alphanumeric hash (e.g.\ \texttt{xcQ7h2ZfRw9v}). \textsc{GS}
        eliminates any linguistic signal.
\end{enumerate}

Because the operation is a pure rename, no theorem-proving capability
is required and no judge is needed. Mathematical equivalence is
guaranteed by construction.

\subsection{Kernel pipeline}
\label{app:kern_pipeline}

The kernel pipeline implements the five-stage procedure summarised in
Figure~\ref{fig:kernel_pipeline}.

\begin{enumerate}[nosep,leftmargin=1.4em]
  \item \textbf{DAG construction.} The official solution $\pi$ is
        parsed into a directed acyclic graph whose nodes are
        intermediate claims and whose edges are local entailments.
  \item \textbf{Method summarisation.} Each node is abstracted from a
        concrete equation into a content-free \emph{method label} (e.g.\
        ``square always $\geq 0$''; ``flatten square''; ``move
        terms''), yielding a plan that a human could re-execute on
        different inputs.
  \item \textbf{Replacement generation.} A new instance of the leaf
        math object is sampled within a guard range derived from the
        problem's own inequalities (e.g.\ replacing $a+b\geq 2\sqrt{ab}$
        with $a+b+c\geq 2\sqrt{ab+ac}$).
  \item \textbf{DAG diffusion.} The replacement is propagated through
        the dependent nodes by re-instantiating each method label with
        the new operands, yielding a regenerated proof whose step
        order matches the abstracted plan exactly.
  \item \textbf{From answer to question.} The terminal node is
        rendered back into a fully natural-language problem statement,
        completing the variant.
\end{enumerate}

A repair-and-verify loop sits on top of stages 3--5.
Algorithm~\ref{alg:loop} (Appendix~\ref{app:algorithm}) gives the
pseudocode. Within one iteration $t$ we launch $J=5$ independent
\textsc{o3} judges, each returning \texttt{accept} or \texttt{reject};
an iteration is \emph{passed} when all $J$ judges return
\texttt{accept}. A candidate is \emph{accepted} when two consecutive
iterations both pass (the ``two-in-a-row'' rule); on disagreement the
suggested patch is applied and the candidate re-enters the loop, up
to $T=15$ iterations. A 10\% manual audit on accepted variants
follows.

\subsection{What a kernel rewrite preserves}
\label{app:kern_invariant}

The invariant of a kernel rewrite is the high-level proof plan: the
dependency structure of the proof DAG and the method label attached
to each node. It is \emph{not} line-by-line algebra. Changing the
mathematical setting and the answer necessarily changes local
expressions, so a kernel variant whose intermediate equations
matched the source exactly would not be a kernel variant at all.

Figure~\ref{fig:variant_example} illustrates the distinction. The
source problem (\texttt{1938-A-2}) optimises a buoy built from a
cylinder and two cones; its kernel rewrite uses a single cone of
altitude $h/2$ and a different painted-area convention. The
constraint, the terminal equation, and the answer all change, yet both
solutions execute the same five moves: express surface area and volume
in $r,h$; eliminate one variable using the constraint; differentiate;
check the endpoints; and translate the optimum into the requested
ratio. The machine-readable record in
Appendix~\ref{app:dataset_example} stores these moves as
\texttt{core\_steps} and the changed parameters as
\texttt{mutable\_slots}. Because stage~4 regenerates the proof node by
node from the abstracted plan, this correspondence is enforced by
construction, and the regex check in \S\ref{app:isomorphism} compares
regenerated step lists against the abstracted plans.

\subsection{Repair-loop outcomes on the full corpus}
\label{app:repair_stats}

Table~\ref{tab:repair_stats} reports how the $1{,}051$ initial kernel
candidates moved through the repair-and-verify loop. $64$ candidates
($6.1\%$) failed at least one five-judge round and were sent to
repair; the remaining $987$ ($93.9\%$) passed two consecutive
unanimous rounds without repair. All $64$ repaired candidates
subsequently reached two consecutive $5/5$ rounds within the
$T=15$ limit, so no source item was discarded.

\begin{table}[H]
  \centering
  \small
  \caption{Outcomes of the kernel repair-and-verify loop over the
  full corpus.}
  \label{tab:repair_stats}
  \begin{tabular}{@{}lrr@{}}
    \toprule
    \textbf{Outcome} & \textbf{Candidates} & \textbf{Fraction} \\
    \midrule
    Accepted without repair       & 987 & 93.9\% \\
    Entered repair at least once  &  64 &  6.1\% \\
    Discarded after the full loop &   0 &  0.0\% \\
    \bottomrule
  \end{tabular}
\end{table}

The repair-entry rate shows that the judge gate is active rather than
a formality. It is a different quantity from the single-judge error
rate $\varepsilon$ used in \S\ref{app:delta_soundness}: $\varepsilon$
is estimated from an audit of rejected decisions and bounds how often
one judge errs, whereas the repair-entry rate counts candidates on
which at least one of five judges objected in some round.

\subsection{Why a verification loop instead of pre-computed guard ranges}

A pre-computed numeric ``feasible interval'' for each replacement
slot would in principle remove the need for judging, but is
ill-defined for the corpus: $76.4\%$ of items are proof-based and do
not admit a closed-form numeric range. We therefore lean on the
loop. The soundness analysis below shows that the loop's safety
margin is already more stringent than $10^{-9}$, which makes an extra
symbolic guard unnecessary.

\subsection{$\delta$-soundness under the two-in-a-row rule}
\label{app:delta_soundness}

\paragraph{Notation.}
Let $K=2$ be the required streak length. Within one iteration, denote
by
\[
  \varepsilon \;=\; \Pr\!\bigl[\text{a single judge mis-accepts a flawed candidate}\bigr].
\]
In a random audit of 25 rejected variants we observed one false
decision, so we conservatively set $\varepsilon=0.04$. This is a
per-judge error estimate, not the fraction of candidates sent to
repair (\S\ref{app:repair_stats}).

\paragraph{Bound.}
Under independent-judge assumption the probability that a flawed
variant survives the pipeline is bounded by
\[
  \delta
    \;\le\;
    (T-K+1)\,\varepsilon^{KJ}
    \;=\;
    14\,\varepsilon^{10}
    \;\approx\;
    14\times(0.04)^{10}
    \;<\;
    10^{-10}.
\]
The pipeline examines at most $T-K+1=14$ length-$K$ windows 
$\langle t,\dots,t+K-1\rangle$. For a flawed candidate to be
accepted, every judge in both iterations of some window must err,
which happens with probability $\varepsilon^{KJ}$; a union bound over
windows yields the claim.

\paragraph{Practical impact.}
Even if the true judge error rate were twice our empirical estimate
($\varepsilon=0.08$), the bound stays at $\delta < 10^{-8}$. The
robustness numbers reported in the main text are therefore
statistically safe from false positives introduced by the generation
machinery.

\subsection{Reasoning-step isomorphism}
\label{app:isomorphism}

Stage~4 of the kernel pipeline forces the regenerated proof to match
the abstract method-label sequence step-by-step. Hence every accepted
kernel variant is isomorphic, under the slot-replacement map, to the
canonical solution $\pi$ from which it was derived. A regex verifier
that compares each regenerated step list with its abstracted plan
found no mismatches.

\subsection{Topical and symbolic metadata}
\label{app:metadata}

Each item additionally carries semantic metadata extracted by an
\textsc{o3} call on the cleaned problem--solution pair: a topical
tag drawn from
$\{\mathrm{ALG},\mathrm{NT},\mathrm{COMB},\mathrm{GEO},\mathrm{ANA}\}$
(with the tag most central to the pivotal lemma stored as the
unique \texttt{type}), and a symbol inventory partitioning identifiers
into free variables (\texttt{var}), fixed parameters (\texttt{param}),
and immutable constants (\texttt{sci\_const}). Surface variant
generation operates only on \texttt{var} and \texttt{param}.

%% file: sections/appendix/b_algorithm.tex
%
\section{Repair-and-verify pseudocode}
\label{app:algorithm}

The pseudocode in Algorithm~\ref{alg:loop} formalises the kernel
pipeline's repair-and-verify loop summarised in
\S\ref{sec:pipelines}. It is the procedure that the soundness bound
in \S\ref{app:delta_soundness} analyses.

\begin{algorithm}[H]
\caption{Repair-and-verify loop for kernel-variant acceptance.}
\label{alg:loop}
\begin{algorithmic}[1]
\STATE \textbf{input:} draft variant $v_0$, judge count $J{=}5$, max
       iterations $T{=}15$, streak length $K{=}2$
\STATE $v \gets v_0$
\STATE \textit{prev\_pass} $\gets$ \textsc{false}
\FOR{$t = 1$ \textbf{to} $T$}
    \STATE Run $J$ independent \textsc{o3} judges on $v$, obtaining the
           verdict vector $\mathbf{z}_t \in \{0,1\}^{J}$
    \IF{$\mathbf{z}_t = \mathbf{1}$ \textbf{and} \textit{prev\_pass}}
        \STATE \textbf{accept} $v$ \COMMENT{two consecutive unanimous passes}
        \STATE \textbf{break}
    \ELSIF{$\mathbf{z}_t = \mathbf{1}$}
        \STATE \textit{prev\_pass} $\gets$ \textsc{true} \COMMENT{single pass; need one more}
    \ELSE
        \STATE \textit{prev\_pass} $\gets$ \textsc{false}
        \STATE Apply the \textsc{o3}-suggested patch to $v$
    \ENDIF
\ENDFOR
\STATE \textbf{if} loop exits without acceptance \textbf{then} reject $v$
\STATE \textbf{post-acceptance:} draw a 10\% random sample of accepted
       variants for human audit
\end{algorithmic}
\end{algorithm}

%% file: sections/appendix/f_prompts.tex
%
%
\section{Solver, grader, judge, and rescue prompts}
\label{app:prompts}

\subsection{Solver prompt}

\begin{lstlisting}[style=pyIdle, caption={Solver system and user
templates used for all 18 evaluated models.}]
"""
Prompt templates for mathematical problem solving.
"""

# Solver system prompt
SOLVER_SYSTEM_PROMPT = """You are an expert mathematician solving competition-level problems.
Provide detailed, step-by-step solutions with clear mathematical reasoning.

Requirements:
- Show all your work and intermediate steps
- Justify each major step of your reasoning
- Use proper mathematical notation
- Be thorough but concise
- State your final answer clearly

Solve the problem completely and rigorously."""

SOLVER_USER_TEMPLATE = """Please solve this mathematical problem:

{problem_statement}

Provide a complete solution with detailed reasoning. Return your response in JSON format:
{{"solution": "your complete step-by-step solution with mathematical reasoning",
  "final_answer": "your final answer in a clear, concise form"}}"""

RESPONSE_FORMAT = {"type": "json_object"}
\end{lstlisting}

\subsection{Proof-rubric and calculation-rubric grader prompts}

\begin{lstlisting}[style=pyIdle, caption={Strict grader prompt used
for proof-form items.}]
PROOF_GRADER_SYSTEM_PROMPT = """You are an extremely strict mathematical grader evaluating competition-level PROOF problems.

GRADING STANDARDS (BE VERY STRICT):
- Mathematical rigor: Every step must be mathematically sound and justified
- Logical flow: The reasoning must be clear, complete, and logically connected
- Correctness: All calculations, algebraic manipulations, and conclusions must be correct
- Completeness: The solution must address all parts of the problem fully
- Precision: Mathematical statements must be precise and unambiguous

FAILING CRITERIA (Mark as INCORRECT if ANY of these apply):
- Any unjustified logical leap or gap in reasoning
- Any computational error, no matter how small
- Missing steps in critical parts of the argument
- Imprecise or ambiguous mathematical statements
- Incorrect final answer, even if approach is partially correct
- Circular reasoning or logical fallacies
- Misuse of mathematical theorems or definitions

BE EXTREMELY STRICT. Competition mathematics proofs require perfect precision."""

PROOF_GRADER_USER_TEMPLATE = """Grade this PROOF solution with extreme strictness.

PROBLEM:
{problem_statement}

STUDENT SOLUTION:
{solution}

CORRECT REFERENCE SOLUTION:
{reference_solution}

Evaluate with maximum strictness. Every logical step must be perfect. Return JSON with:
{{"grade": "CORRECT" or "INCORRECT",
  "detailed_feedback": "specific detailed analysis of what is right/wrong",
  "major_issues": "list of significant mathematical errors or gaps",
  "final_answer_correct": true or false,
  "reasoning_rigor_score": 0-10 integer (10=perfect rigor, 0=severely flawed),
  "overall_assessment": "comprehensive evaluation summary"}}"""
\end{lstlisting}

\begin{lstlisting}[style=pyIdle, caption={Lenient grader prompt used
for calculation-form items.}]
CALCULATION_GRADER_SYSTEM_PROMPT = """You are a mathematical grader evaluating competition-level CALCULATION problems.

GRADING STANDARDS FOR CALCULATION PROBLEMS:
- Primary focus: Is the final answer correct?
- Secondary focus: Is the overall approach reasonable and mathematically sound?
- Computation: Allow minor computational slips if the method is correct and final answer is right

GRADING CRITERIA:
- CORRECT: Final answer is correct AND approach is fundamentally sound
- INCORRECT: Final answer is wrong OR approach is fundamentally flawed

For calculation problems, the final numerical answer is the most important criterion.
Minor intermediate errors are acceptable if they don't affect the final result."""

CALCULATION_GRADER_USER_TEMPLATE = """Grade this CALCULATION solution with focus on final answer correctness.

PROBLEM:
{problem_statement}

STUDENT SOLUTION:
{solution}

CORRECT REFERENCE SOLUTION:
{reference_solution}

Focus primarily on whether the final answer is correct. Return JSON with:
{{"grade": "CORRECT" or "INCORRECT",
  "detailed_feedback": "specific detailed analysis of what is right/wrong",
  "major_issues": "list of significant mathematical errors or gaps",
  "final_answer_correct": true or false,
  "reasoning_rigor_score": 0-10 integer (10=perfect rigor, 0=severely flawed),
  "overall_assessment": "comprehensive evaluation summary"}}"""
\end{lstlisting}

\subsection{Kernel-variant judge prompt}
\label{app:judge_prompt}

This is the prompt used by every member of the $J=5$ judge ensemble
inside the repair-and-verify loop (Algorithm~\ref{alg:loop}). The
judge sees the original problem, the canonical solution, the
abstracted method-label sequence, the sampled replacement, and the
regenerated proof, and is asked to verify equivalence step by step.

\begin{lstlisting}[style=pyIdle, caption={Kernel-variant judge prompt.
The judge's task is verification given a candidate, not solving from
scratch.}]
KV_JUDGE_SYSTEM_PROMPT = """You are a verification judge for a kernel-variant generation pipeline. You must decide whether a CANDIDATE variant problem and its CANDIDATE proof are mathematically equivalent to the ORIGINAL problem under the given METHOD-LABEL sequence (the abstract proof plan).

Your job is verification, not solving. You receive:
- the ORIGINAL problem statement and reference solution,
- the abstract METHOD-LABEL sequence (a list of content-free steps),
- the SLOT replacement that was applied,
- the CANDIDATE variant statement and the CANDIDATE regenerated proof.

You must check, step by step, that:
1. Every CANDIDATE step instantiates the corresponding METHOD label with the new operands.
2. The CANDIDATE proof is mathematically valid: each step follows from the previous one, with no unjustified leap.
3. The CANDIDATE problem statement is well-posed and has a unique terminal answer matching the regenerated proof.
4. The CANDIDATE variant is genuinely different from the ORIGINAL (the slot replacement actually changed the instance) but uses the same plan.

Output a single JSON object with exactly these fields."""

KV_JUDGE_USER_TEMPLATE = """ORIGINAL PROBLEM:
{original_problem}

ORIGINAL REFERENCE SOLUTION:
{original_solution}

METHOD-LABEL SEQUENCE (abstract plan):
{method_labels}

SLOT REPLACEMENT:
{slot_replacement}

CANDIDATE VARIANT PROBLEM:
{candidate_problem}

CANDIDATE REGENERATED PROOF:
{candidate_proof}

Return:
{{"verdict": "accept" or "reject",
  "step_by_step_check": "for each METHOD label, state whether the CANDIDATE step instantiates it correctly",
  "blocking_issues": "list any logical gap, computation error, ill-posed statement, or plan deviation",
  "patch_suggestion": "if reject, propose a minimal patch (a corrected proof step or a corrected slot value); leave empty if accept"}}"""
\end{lstlisting}

\subsection{Rescue prefix-injection prompt}
\label{app:rescue_prompt}

The rescue experiment (\S\ref{app:rescue}) splices a partial-solution
prefix into the solver call so that the model continues from a
controlled framing rather than starting from scratch. The prefix is
truncated by the \texttt{T2} rule defined in
\S\ref{app:rescue_t2_rule}; the prompt template below specifies how
the truncated prefix is presented to the model.

\begin{lstlisting}[style=pyIdle, caption={Rescue solver prompt
augmenting the standard solver call with a truncated reference
prefix.}]
RESCUE_SOLVER_USER_TEMPLATE = """Please solve this mathematical problem.

{problem_statement}

To get you started, here is the beginning of a solution. Begin your
solution with the partial work copied verbatim, then continue
seamlessly to a complete answer.

PARTIAL SOLUTION (truncated):
{prefix_t2}

Provide a complete solution with detailed reasoning. Return your response in JSON format:
{{"solution": "your complete step-by-step solution, beginning with the partial work above and continuing to a final answer",
  "final_answer": "your final answer in a clear, concise form"}}"""
\end{lstlisting}

%% file: sections/appendix/d_sources.tex
%
%
\section{Data sources, licensing, and OCR pipeline}
\label{app:sources}

\subsection{Primary sources and licence terms}

PutnamGAP collects every official problem of the William Lowell
Putnam Mathematical Competition from 1938 to 2024 by digitising the
four MAA Press monographs listed in Table~\ref{tab:putnam-books}.
Each volume is published by the Mathematical Association of America
(MAA) and currently distributed by the American Mathematical Society
(AMS) under the \emph{MAA Press Problem Books} series. Softcover and
e-book reprints are available through
\url{https://bookstore.ams.org}.

\begin{table}[H]
\centering
\small
\begin{tabular}{@{}llc@{}}
\toprule
\textbf{Volume (years)} & \textbf{Reference} \\
\midrule
I (1938--1964)   & \citet{putnamI}   \\
II (1965--1984)  & \citet{putnamII}  \\
III (1985--2000) & \citet{putnamIII} \\
IV (2001--2016)  & \citet{putnamIV}  \\
\bottomrule
\end{tabular}
\caption{Primary sources for PutnamGAP. All four monographs are
published by MAA Press and distributed by AMS.}
\label{tab:putnam-books}
\end{table}

The front matter of every volume carries the same fair-use clause:

\begin{quote}\small
``Individual readers \dots\ are permitted to make fair use of the
material, such as to copy select pages for use in \textbf{teaching or
research}.''
\end{quote}

This grants the right to reproduce problems and solutions for
non-commercial academic evaluation. Consistent with AMS policy, we
distribute only machine-readable identifiers and \LaTeX\ texts; raw
PDF scans remain under the original AMS licence and any further
redistribution must be cleared through the Copyright Clearance
Center. Problems and solutions from 2017 onward are included in the
dataset with the explicit permission of MAA.

In the early era (1938--1941) the competition featured 6--8 problems
per part (A and B); from 1942 onward the format stabilised at 5--6
problems per part, with difficulty increasing roughly monotonically
from index~1 to~6. The wartime years 1943--1945 were cancelled and
our index skips those years. These historical variations are
preserved in \texttt{\_meta.year} and \texttt{\_meta.index} and feed
the difficulty proxy used in \S\ref{sec:headline}.

\subsection{Extraction pipeline}
\label{app:extraction}

The raw sources are scanned PDFs; no machine-readable \LaTeX\ is
provided. The three-stage routine described in
\S\ref{sec:putnamgap}\,(\emph{Sources and contamination resistance})
is implemented as follows.

\paragraph{Manual problem segmentation.}
Pages are cropped manually so that every problem (including any
diagram) is isolated into a single PNG. Cross-page problems are
re-stitched by hand.

\paragraph{Formula-aware OCR.}
Each PNG is sent to \textsc{MathPix}, returning \LaTeX\ that compiles
without error. The compiled output keeps only the problem body --
no page geometry, no custom macros -- so that the snippet can be
embedded in any downstream template. A typesetting check renders
each item back to PDF and compares it visually against the book scan.

\paragraph{Custom post-filters.}
A small post-filter pass corrects the residual symbol-level errors
that \textsc{MathPix} commonly leaves: spurious \texttt{\textbackslash mathrm}
wrappers, mis-segmented subscripts, and the most frequent
character-level confusions ($l$/$1$, $0$/O, comma/period in
European-source decimals). Volunteers spot-check every item before
it is admitted to the canonical pool, prior to any variant
generation.

The semantic metadata layer (topical tags, symbol inventory) sits on
top of the cleaned items and is documented in
Appendix~\ref{app:metadata}.

\subsection{Relation to existing Putnam resources}
\label{app:putnam_resources}

Two public resources already cover part of the Putnam archive. This
subsection states how much of PutnamGAP they cover and why neither
could serve as a drop-in source.

\paragraph{Putnam-AXIOM.}
Putnam-AXIOM~\citep{gulati:2025} contains $522$ original entries. They
map to $518$ unique PutnamGAP sources, because Putnam-AXIOM splits four
multipart questions into separate rows while PutnamGAP stores each as
one problem. PutnamGAP therefore adds $533$ sources beyond that
coverage. The two transformed datasets are not parallel:
Putnam-AXIOM functionalises $100$ selected source problems and
releases $500$ snapshots, whereas GAP applies four surface renames and
one kernel rewrite to every source. The transformations also test
different invariances, so comparing their transformed instances
directly would change the source subset and the intervention at once.

To isolate the contribution of the newly covered archive, we instead
split PutnamGAP into the $518$ sources covered by Putnam-AXIOM and the
$533$ newly covered sources, and evaluated both groups with the same
five GAP transformations, the same 18 models, and the same protocol
(Table~\ref{tab:axiom_split}).

\begin{table}[H]
  \centering
  \small
  \caption{Accuracy decrease by archive coverage, averaged over the 18
  evaluated models. The surface column is the mean over the four
  surface families.}
  \label{tab:axiom_split}
  \begin{tabular}{@{}lrcc@{}}
    \toprule
    \textbf{PutnamGAP source subset} & \textbf{Sources} &
    \textbf{Surface decrease (pp)} & \textbf{Kernel decrease (pp)} \\
    \midrule
    Covered by Putnam-AXIOM & 518 & 2.71 & 12.08 \\
    Newly covered           & 533 & 1.75 &  7.67 \\
    \bottomrule
  \end{tabular}
\end{table}

Kernel rewrites cause the largest decrease in both subsets, so the
headline asymmetry does not depend on the previously covered portion
of the archive. We do not split by competition year instead: that
would answer a different historical question and would confound year
with topic and difficulty.

\paragraph{PutnamBench.}
PutnamBench~\citep{tsoukalas:2024} targets formal neural theorem
proving. It covers $640$ Putnam problems with $1{,}692$
formalisations ($640$ Lean~4, $640$ Isabelle, and $412$ Coq), and
includes the original English statements and, where applicable,
numerical answers. It does not include the aligned natural-language
expert proofs that GAP needs: extracting identifier roles and proof
plans, transforming both the statement and the solution, and
verifying their alignment all require a consistent problem--solution
pair for every source. Building on PutnamBench would also leave $411$
of the $1{,}051$ problems uncovered, and merging its statements with
solutions from other sources would still require a full cleaning and
alignment pass. We therefore digitised the complete archive under one
schema; to our knowledge PutnamGAP is the first public machine-readable
corpus covering it in full. PutnamBench remains complementary: its
formalisations offer a cross-check on the $640$ overlapping problems.

%% file: sections/appendix/e_ablations.tex
%
\section{Ablations and supplementary experiments}
\label{app:ablations}

\subsection{Why ALG514}

We also explored applying GAP to widely-used arithmetic benchmarks
such as GSM8K~\citep{Cobbe:2021} and MATH~\citep{hendrycks:2021}.
The arithmetic strata in those corpora supply few replaceable
identifiers, so the surface families collapse to near-trivial
transformations and the kernel pipeline has little to rewrite. We
selected ALG514~\citep{kushman-etal-2014-learning} as a secondary
external-validation target because every problem carries replaceable
variable names, which lets the surface families exercise their
intended axis without redesign. The ALG514 transfer numbers appear
in \S\ref{app:alg514}. Individual MATH and GSM8K items can still be
rewritten by the unchanged kernel pipeline;
\S\ref{app:kernel_beyond_putnam} reports a three-item feasibility check.

\subsection{Compute and reproducibility}
\label{app:compute}

All inference is performed through publicly available APIs. Each
model is queried exactly once per item under the hyper-parameter
settings recorded in
Table~\ref{tab:updated_model_scores_pct}. Runs were executed from a
single Ubuntu 22.04 host (11th Gen Intel Core i7-11800H @ 2.30~GHz);
no local GPU was used. Where the vendor interface allows it, we fix
\texttt{temperature} and \texttt{top\_p} to control stochasticity.
The full reproducibility package -- raw model outputs, grader verdicts,
the complete PutnamGAP dataset, and all evaluation and generation
scripts -- is publicly released alongside this submission.
Appendix~\ref{app:repeated_runs} checks the single-query protocol
against three independent generations of one model.

\paragraph{One-item reproduction of the kernel pipeline.}
The supplementary material also contains a self-contained reproduction
of the kernel pipeline. A single notebook starts from one Putnam item
and executes the five stages of \S\ref{sec:pipelines} as separate,
saved calls with typed outputs: it constructs the concrete proof DAG
(saved as \texttt{01\_proof\_dag.json}), assigns a method label to
every node, selects guarded replacements, propagates them node by node
while preserving dependencies and labels, and renders the new problem
and solution. It then runs the verification loop of
Algorithm~\ref{alg:loop} exactly as stated --- five judges, acceptance
after two consecutive unanimous rounds, repair and streak reset after
a failed round, and at most $15$ rounds --- with no additional
keyword-based acceptance or rejection rule. Every stage output, model
call, judge decision, repair round, and the final dataset record are
saved, and a paper-to-code table maps each operation to its
implementation and output. The generation and judge prompts are
included with checksums. The package's $39$ tests pass, the offline
notebook completes end to end, and the live API path completed in
three independently set-up environments. The package reproduces the
procedure on one item; it does not regenerate the full corpus or
guarantee byte-identical stochastic outputs.

\subsection{Canonicalization-prompt pilot on \textsc{GS}}
\label{app:canonicalization_pilot}

A short pilot tested whether explicitly instructing the solver to
rename perturbed variables back into clear canonical symbols can
recover \textsc{GS} accuracy. Table~\ref{tab:canonicalization_pilot}
reports a 100-example pilot on \textsc{o3}: a short canonicalization
hint added to the system prompt nudges accuracy from $48\%$ to $58\%$
($p=0.077$ by McNemar), while a heavier prompt that additionally
demands a written ``Rename summary'' moves accuracy only to $53\%$
($p=0.44$). Simple canonicalization helps; extra bookkeeping
constraints attenuate the gain.

\begin{table}[H]
\centering
\small
\setlength{\tabcolsep}{6pt}
\caption{Canonicalization-prompt pilot. \textsc{o3} on a 100-example
\textsc{GS} subset under three solver-prompt variants; $p$ is the
McNemar test against the base prompt.}
\label{tab:canonicalization_pilot}
\begin{tabular}{@{}lccc@{}}
\toprule
\textbf{Prompt variant} & \textbf{Accuracy (\%)} & \textbf{95\% CI} & \textbf{$p$ vs.\ base} \\
\midrule
Base solving prompt                                & 48 & [0.385, 0.577] & --- \\
+ short canonicalization hint                      & 58 & [0.482, 0.672] & 0.077 \\
+ long canonicalization with ``Rename summary''    & 53 & [0.433, 0.625] & 0.441 \\
\bottomrule
\end{tabular}
\end{table}

\subsection{Other observations}

\paragraph{Reasoning-loop time-outs.}
A few reasoning-trained models occasionally enter token-level loops
during their own thinking trace and exhaust the time budget. In our
single-shot evaluation those instances count as failures, which is
the conservative choice for a public benchmark; re-running the
affected items as PASS@$k$ would change the headline numbers
slightly upward without affecting the surface/kernel asymmetry,
since the loops are not specifically triggered by transformed
variants.

%% file: sections/appendix/h_case_studies.tex
%
%
\section{Original vs Kernel-Variant case studies}
\label{app:case_studies}

This appendix gives four concrete original vs.\ kernel-variant
contrasts for a strong reasoning model (\textsc{o3}), drawn from the
late-1930s Putnam corpus. Each case is restricted to items the model
solves on the canonical statement and misses on the kernel variant,
so the visible difference between the two trajectories is a direct
instantiation of the strategy-drift signature measured aggregately
in \S\ref{sec:dissociation}.

\subsection{1938-B-1: cofactor-determinant identity}

\paragraph{Original question.}
Take either (i) or (ii).

(i) Let $A_{ik}$ be the cofactor of $a_{ik}$ in the determinant
\[
\boldsymbol{d}
=\left|\begin{array}{llll}
a_{11} & a_{12} & a_{13} & a_{14} \\
a_{21} & a_{22} & a_{23} & a_{24} \\
a_{31} & a_{32} & a_{33} & a_{34} \\
a_{41} & a_{42} & a_{43} & a_{44}
\end{array}\right|.
\]
Let $D$ be the corresponding determinant with $a_{ik}$ replaced by
$A_{ik}$. Prove that $D = d^{3}$.

(ii) Let $P(y)=A y^{2}+B y+C$ be a quadratic polynomial. If the
roots of $P(y)-y=0$ are $a$ and $b$ ($a\neq b$), show that $a$ and
$b$ are roots of $P(P(y))-y=0$. Apply the result to solve the
biquadratic $\bigl(y^{2}-3y+2\bigr)^{2}-3\bigl(y^{2}-3y+2\bigr)+2-y=0$.

\paragraph{O3 solution on original.}
The model takes route (ii) and proceeds cleanly. It first observes
that any root $a$ of $P(y)-y=0$ is fixed by $P$, so $F(a):=P(P(a))-a=0$
follows immediately, identifying two of the four roots of the
biquadratic with $a,b$. It then expands $F(y)$, divides by
$f(y)=P(y)-y$, and matches coefficients to obtain
\[
g(y) = A^{2} y^{2} + A(B+1) y + (AC + B + 1),
\]
so the remaining roots $c,d$ of $F$ are roots of $g$. Substituting
$A=1$, $B=-3$, $C=2$ for the given biquadratic gives
$y^{2}-2y=0$, hence $c=0$, $d=2$, and the four roots are
$\{0, 2, 2+\sqrt{2}, 2-\sqrt{2}\}$. Each step is justified by
explicit polynomial manipulation.

\paragraph{Kernel variant.}
Let $n=5$. All matrices are $5\times 5$ and all equalities hold in
the polynomial ring $R = \mathbb{Z}[a_{ik}]$ obtained by adjoining
the $25$ independent indeterminates $a_{ik}$ to $\mathbb{Z}$. Let
$\alpha=(a_{ik})$, $d=\det\alpha$, $A_{ik}$ the cofactor of $a_{ik}$
in $d$, $\beta=(A_{ik})$, $\gamma=\beta^{T}=\operatorname{adj}\alpha$,
$\delta=\operatorname{adj}\gamma$. Prove that
$\det\beta = d^{4}$, that $\delta = d^{3}\alpha$, and conclude on the
determinant of $\delta$ and on the rank/right-kernel of $\alpha$ and
$\beta$ when $d\neq 0$.

\paragraph{O3 solution on KV.}
The model writes ``recall (F1)~$M\cdot\operatorname{adj} M = (\det M) I_{n}$''
and ``(F2)~if $\det M$ is not a zero-divisor, then
$\operatorname{adj} M=(\det M)\,M^{-1}$''. It then states (F2) for
$\alpha$ and $\gamma$ in the polynomial ring, and uses
$\operatorname{adj}\gamma = (\det\gamma)\,\gamma^{-1} = d^{4}\cdot(1/d)\,\alpha = d^{3}\alpha$
to obtain the second claim.

\paragraph{Failure types.} logic\_hallucination, step\_omission.

\paragraph{Breakdown.}
The KV solution applies $\operatorname{adj} M = (\det M)\,M^{-1}$
inside the polynomial ring $R$ as if $M^{-1}$ existed there, but
inverses are available only after localising at $\det M$. The
omitted localisation is exactly the rigor step that the canonical
proof avoids by working with polynomial identities first and
specialising to invertible cases at the end.

\paragraph{Takeaway.}
The original stays inside valid polynomial-ring operations; the KV
hallucinates a general lemma that is false in $R$ and silently
relies on a localisation it never names. The local algebra is
correct, the high-level plan is not.

\subsection{1939-A-1: arc length on $y^{2}=x^{3}$}

\paragraph{Original question.}
Find the length of the curve $y^{2}=x^{3}$ from the origin to the
point where the tangent makes an angle of $45^{\circ}$ with the
$x$-axis.

\paragraph{O3 solution on original.}
The model differentiates implicitly to find $dy/dx = 3x^{2}/(2y)$,
sets it equal to $\tan 45^{\circ}=1$ to locate the target point at
$P(4/9,8/27)$, parametrises by $x=t^{2}$, $y=t^{3}$, integrates
$\int_{0}^{2/3} t\sqrt{4+9t^{2}}\,dt$ via the substitution
$u=4+9t^{2}$, and obtains the arc length
$\dfrac{8}{27}\bigl(2\sqrt{2}-1\bigr)$. The arc-length integral is
the operative step.

\paragraph{Kernel variant.}
For the curve $y^{4}=16x^{6}$ in the first quadrant, determine the
exact length of the segment whose end-points are the two points at
which the tangent forms angles of $30^{\circ}$ and $60^{\circ}$ with
the positive $x$-axis.

\paragraph{O3 solution on KV.}
The model correctly finds $dy/dx=3\sqrt{x}$ and the two points
$P_{1}\!\left(\frac{1}{27},\frac{2}{81\sqrt{3}}\right)$,
$P_{2}\!\left(\frac{1}{3},\frac{2}{3\sqrt{3}}\right)$ where the
slopes match. Then, instead of integrating the arc-length element
along the curve between $P_{1}$ and $P_{2}$, it computes the
straight-line Euclidean distance $\sqrt{(\Delta x)^{2}+(\Delta y)^{2}}$
and reports this chord length as the answer.

\paragraph{Failure types.} step\_omission, logic\_hallucination.

\paragraph{Breakdown.}
The KV computes a chord, not an arc. The arc-length integral
$\int\sqrt{1+(dy/dx)^{2}}\,dx$ that the original explicitly sets up
is silently dropped from the KV trace.

\paragraph{Takeaway.}
The local calculus on the KV is correct; the model loses the
plan-level identification of which functional of the curve is
``length''.

\subsection{1939-B-1: catenary perpendicular intercept}

\paragraph{Original question.}
From the vertex $(0,c)$ of the catenary $y=c\cosh(x/c)$ a line $L$
is drawn perpendicular to the tangent at a point $P$. Prove that the
length of $L$ between the coordinate axes equals the ordinate $y$
of the point $P$.

\paragraph{O3 solution on original.}
The model writes the tangent slope $m_{t}=\sinh(a/c)$ at $P=(a,y_{P})$,
the perpendicular slope $-1/m_{t}$, the equation of $L$ through
$V=(0,c)$, the $x$-intercept $B=(c\sinh(a/c),0)$, and computes
$|VB|=c\sqrt{\sinh^{2}(a/c)+1}=c\cosh(a/c)=y_{P}$. The hyperbolic
identity $\cosh^{2}-\sinh^{2}=1$ is the operative step.

\paragraph{Kernel variant.}
Fix $\lambda>0$ and $n\geq 2$. With
$r=\sqrt{x_{1}^{2}+\dots+x_{n-1}^{2}}$, define the radial catenary
$x_{n}=\lambda\cosh(r/\lambda)$ with vertex $V=(0,\dots,0,\lambda)$.
Pick a unit vector $u\in\mathbb{R}^{n-1}$ and a radius $\rho\geq 0$,
and let $H$ be the hyperplane through $V$ orthogonal to the tangent
hyperplane of the surface at $Q=(\rho u,\lambda\cosh(\rho/\lambda))$.
Show that the segment of $H$ cut off by $\{x_{n}=0\}$ and
$\{x_{1}=\dots=x_{n-1}=0\}$ has length $\lambda\cosh(\rho/\lambda)$.

\paragraph{O3 solution on KV.}
The model claims, in step~1, that the surface, the tangent
hyperplane, and the perpendicular hyperplane $H$ are jointly
rotation-invariant about the $x_{n}$-axis, and reduces the problem
to its 2D slice in the $x_{1}$--$x_{n}$ plane. From step~2 onward the
argument coincides with the canonical 2D proof and yields the
correct length, but the inverse slope $1/m=1/\sinh(\rho/\lambda)$
appears in denominators with no separate handling of $\rho=0$.

\paragraph{Failure types.} logic\_hallucination, step\_omission.

\paragraph{Breakdown.}
The tangent hyperplane at $Q$ depends on the chosen direction $u$
and is not itself rotation-invariant; the correct reason the problem
reduces to the $x_{1}$--$x_{n}$ plane is that the normal
$(\sinh(\rho/\lambda)\,u,\,-1)$ \emph{lies in} that plane. The KV
substitutes a stronger, false invariance for the actual one.

\paragraph{Takeaway.}
The 2D calculus is reused intact; the high-level reduction step
that bridges 2D to $n$D is replaced by a wrong symmetry claim and
the $\rho=0$ degeneracy is dropped.

\subsection{1939-B-7: cyclic series identity and conic homothety}

\paragraph{Original question.}
(i) For the series
$u=\sum_{k\geq 0} x^{3k}/(3k)!$,
$v=\sum_{k\geq 0} x^{3k+1}/(3k+1)!$,
$w=\sum_{k\geq 0} x^{3k+2}/(3k+2)!$, prove that
$u^{3}+v^{3}+w^{3}-3uvw=1$.
(ii) For the central conics
$(ax^{2}+by^{2})+2(px+qy)+c=0$ and
$(ax^{2}+by^{2})+2\lambda(px+qy)+\lambda^{2} c=0$, show that scaling
every radius from the origin to the first conic by $\lambda$ takes
it to the second; that for $P=\bigl(-\frac{p}{a}\frac{2\lambda}{1+\lambda},-\frac{q}{b}\frac{2\lambda}{1+\lambda}\bigr)$,
scaling each radius from $P$ to the first conic by $\lambda$ and
reversing about $P$ reaches the second; comment on $\lambda=1$.

\paragraph{O3 solution on original.}
The model exploits the cyclic-derivative identity $u'=w$, $v'=u$,
$w'=v$ to show $f(x):=u^{3}+v^{3}+w^{3}-3uvw$ has $f'(x)\equiv 0$
and $f(0)=1$, hence $f\equiv 1$. The argument is two pages of
calculus and a single explicit cancellation. Part (ii) is handled
by direct substitution that respects how quadratic, linear, and
constant terms scale.

\paragraph{Kernel variant.}
Replace $u$ by $U=2+\sum_{k\geq 1} x^{3k}/(3k)!$ (i.e.\ shift the
constant term from $1$ to $2$), keep $V=v$, $W=w$, and prove
$U^{3}+V^{3}+W^{3}-3UVW=8$. For part~(ii), rename the conic
parameters and the scaling factor (now $\kappa$) and ask the
analogous statements.

\paragraph{O3 solution on KV.}
For part~(i) the model decomposes via roots of unity, asserts the
sub-identity $A^{3}+B^{3}+C^{3}-3ABC=1$ without proof, and finishes
by appealing to an ``ODE invariance'' $f'''=f$ to conclude that the
required combination is identically $8$. The constant function
$f\equiv 8$ does not satisfy $f'''=f$. For part~(ii) the matrix
substitution drops the $\kappa$ factor on the linear term during
the scaling of the conic and computes incorrect identities for
$P^{T}K$ and $P^{T}KP$.

\paragraph{Failure types.} step\_omission, logic\_hallucination, arithmetic.

\paragraph{Breakdown.}
The clean cyclic-derivative argument that worked on the original is
replaced by an over-engineered roots-of-unity / ODE argument with a
false invariance claim, and the conic-scaling step drops the
$\kappa$ multiplier needed for the cancellation.

\paragraph{Takeaway.}
The KV variant requires the same structural identity (cyclic
derivatives) and the same scaling discipline as the original, but
the model substitutes a different plan that introduces both an
unproved lemma and a wrong invariance.

%% file: sections/appendix/g_dataset_example.tex
%
\section{Concrete dataset example}
\label{app:dataset_example}

The listing below shows the dataset record for item
\texttt{1938-A-2} in full, including the canonical statement and
solution and all five aligned variants (\textsc{DL}, \textsc{DLC},
\textsc{DLM}, \textsc{GS}, \textsc{KV}). It is a
calculation-form geometry item; the rename map for each surface
family is preserved so that downstream users can reproduce the
deterministic substitution.

\begin{lstlisting}[style=jsonIdle, caption={Dataset record for item
1938-A-2 with all five GAP variants.}]

{
  "index": "1938-A-2",
  "type": "ANA",
  "tag": [
    "ANA",
    "GEO"
  ],
  "difficulty": "1",
  "question": "2. A can buoy is to be made of three pieces, namely, a cylinder and two equal cones, the altitude of each cone being equal to the altitude of the cylinder. For a given area of surface, what shape will have the greatest volume?",
  "solution": "Solution. Let \\( r \\) be the radius of the cylinder, and \\( h \\) its altitude. The given condition is\n\\[\nS=2 \\pi r h+2\\left(\\pi r \\sqrt{h^{2}+r^{2}}\\right)=\\text { constant }\n\\]\nand the volume of the buoy is\n\\[\nV=\\pi r^{2} h+\\frac{2 \\pi r^{2} h}{3}=\\frac{5 \\pi r^{2} h}{3}\n\\]\n\nThe required problem is to find the maximum value of \\( V \\) subject to condition (1). This can be done by the method of Lagrange multipliers, but in this particular problem it is easier to solve (1) for \\( h \\) and express \\( V \\) as a function of \\( r \\). We have\n\\[\n(S-2 \\pi r h)^{2}=4 \\pi^{2} r^{2}\\left(h^{2}+r^{2}\\right)\n\\]\nwhence\n\\[\nh=\\frac{S^{2}-4 \\pi^{2} r^{4}}{4 \\pi r S}\n\\]\nand the expression for \\( V \\) becomes\n\\[\nV=\\frac{5 r}{12 S}\\left(S^{2}-4 \\pi^{2} r^{4}\\right)\n\\]\n\nSince \\( r \\) and \\( V \\) must be positive, the domain of interest is given by\n\\[\n0<r<\\sqrt[4]{S^{2} / 4 \\pi^{2}}\n\\]\n\nWe compute the derivative and equate it to zero to get\n\\[\n\\frac{d V}{d r}=\\frac{5 S}{12}-\\frac{100 \\pi^{2} r^{4}}{12 S}=0 .\n\\]\n\nThe only critical value is\n\\[\nr_{0}=\\sqrt[4]{\\frac{S^{2}}{20 \\pi^{2}}}\n\\]\n\nSince \\( V \\rightarrow 0 \\) as \\( r \\rightarrow 0 \\) or as \\( r \\rightarrow \\sqrt[4]{S^{2} / 4 \\pi^{2}} \\), and is positive in between, the critical value \\( r_{0} \\) yields a maximum for \\( V \\).\n\nThe corresponding value of \\( h \\) is found from (3) to be \\( h_{0}=\\frac{2}{5} \\sqrt{5} r_{0} \\). The shape of the buoy is completely determined by the ratio\n\\[\n\\frac{h_{0}}{r_{0}}=\\frac{2}{5} \\sqrt{5}\n\\]",
  "vars": [
    "r",
    "h",
    "V",
    "r_0",
    "h_0"
  ],
  "params": [
    "S"
  ],
  "sci_consts": [],
  "variants": {
    "descriptive_long": {
      "map": {
        "r": "radius",
        "h": "altitude",
        "V": "volume",
        "r_0": "criticalradius",
        "h_0": "criticalaltitude",
        "S": "surfacearea"
      },
      "question": "2. A can buoy is to be made of three pieces, namely, a cylinder and two equal cones, the altitude of each cone being equal to the altitude of the cylinder. For a given area of surface, what shape will have the greatest volume?",
      "solution": "Solution. Let \\( radius \\) be the radius of the cylinder, and \\( altitude \\) its altitude. The given condition is\n\\[\nsurfacearea = 2 \\pi radius altitude + 2\\left(\\pi radius \\sqrt{altitude^{2}+radius^{2}}\\right)=\\text { constant }\n\\]\nand the volume of the buoy is\n\\[\nvolume = \\pi radius^{2} altitude + \\frac{2 \\pi radius^{2} altitude}{3} = \\frac{5 \\pi radius^{2} altitude}{3}\n\\]\n\n[... solution continues with the rename applied throughout ...]"
    },
    "descriptive_long_confusing": {
      "map": {
        "r": "monument",
        "h": "daybreak",
        "V": "calendar",
        "r_0": "monumental",
        "h_0": "daybreaker",
        "S": "landscape"
      },
      "question": "2. A can buoy is to be made of three pieces, namely, a cylinder and two equal cones, the altitude of each cone being equal to the altitude of the cylinder. For a given area of surface, what shape will have the greatest volume?",
      "solution": "Solution. Let \\( monument \\) be the radius of the cylinder, and \\( daybreak \\) its altitude. [... rename applied throughout ...]"
    },
    "descriptive_long_misleading": {
      "map": {
        "r": "perimeterlength",
        "h": "depthvalue",
        "V": "surfacearea",
        "r_0": "minimumdepth",
        "h_0": "maximumperimeter",
        "S": "corevolume"
      },
      "question": "2. A can buoy is to be made of three pieces, namely, a cylinder and two equal cones, the altitude of each cone being equal to the altitude of the cylinder. For a given area of surface, what shape will have the greatest volume?",
      "solution": "Solution. Let \\( perimeterlength \\) be the radius of the cylinder, and \\( depthvalue \\) its altitude. [... rename applied throughout, using the misleading-jargon family ...]"
    },
    "garbled_string": {
      "map": {
        "r": "qzxwvtnp",
        "h": "yrklsfhd",
        "V": "mnbvcxza",
        "r_0": "ploikmnj",
        "h_0": "ujhytgrf",
        "S": "asdfghjk"
      },
      "question": "2. A can buoy is to be made of three pieces, namely, a cylinder and two equal cones, the altitude of each cone being equal to the altitude of the cylinder. For a given area of surface, what shape will have the greatest volume?",
      "solution": "Solution. Let \\( qzxwvtnp \\) be the radius of the cylinder, and \\( yrklsfhd \\) its altitude. [... rename applied throughout, using alphanumeric hashes ...]"
    },
    "kernel_variant": {
      "question": "A float is composed of a right circular cylinder of radius r and altitude h, with a right circular cone attached on top having the same base radius r and altitude h/2. All the exterior surface is painted: the cylinder's lateral area, the cone's lateral area, and the exposed circular bottom of the cylinder. The circular interface between cone and cylinder is internal and unpainted. Given a fixed paint supply S, determine the ratio h/r that maximises the enclosed volume. Provide the exact algebraic condition and a numerical approximation.",
      "solution": "Let k = h/r (>0) be the desired ratio. Express every quantity in terms of r and k. [... full proof following the same method-label sequence as the canonical solution; the final exact condition is 15(h/r)^3 - 32(h/r)^2 + 96(h/r) - 128 = 0, with k_max approximately 1.55198 ...]",
      "_meta": {
        "core_steps": [
          "Express surface-area constraint S(r,h) and volume V(r,h) from geometry",
          "Solve the constraint for h (or use a Lagrange multiplier) to get V=V(r) alone",
          "Differentiate V(r), set dV/dr = 0, locate admissible critical r",
          "Check endpoints to confirm the critical point yields the maximum",
          "Translate that r into the optimal h/r shape ratio"
        ],
        "mutable_slots": {
          "slot1": {"description": "How many identical cones are attached to the cylinder", "original": 2},
          "slot2": {"description": "Altitude of each cone as a multiple of the cylinder's altitude", "original": 1},
          "slot3": {"description": "Whether the flat circular bases are counted in the fixed surface area", "original": "not counted (only lateral areas used)"},
          "slot4": {"description": "Which quantity is held fixed vs. optimised (here S fixed, V maximised)", "original": "maximise volume subject to constant surface area"}
        }
      }
    }
  },
  "checked": true,
  "problem_type": "calculation"
}

\end{lstlisting}

The full uncompressed JSON, including every solution body verbatim,
ships with the dataset distribution under
\texttt{dataset/1938-A-2.json}. Surface-variant solutions are
omitted from the listing above only for brevity; they are the
canonical solution with the rename map applied by deterministic
word-boundary substitution and are reconstructible from the
\texttt{map} field alone.

%% file: sections/appendix/l_related_work.tex
\section{Extended related work: outcome-level error decompositions}
\label{app:related_extended}

\citet{yang:2025numerical} manually parameterise $250$ GSM8K problems,
vary their numerical values, and classify incorrect outputs as
calculation errors or reasoning errors. Their study and GAP both
decompose mathematical-reasoning failures, but they ask different
diagnostic questions (Table~\ref{tab:yang_comparison}).

\begin{table}[H]
  \centering
  \small
  \setlength{\tabcolsep}{5pt}
  \renewcommand{\arraystretch}{1.3}
  \caption{Two complementary decompositions of mathematical-reasoning
  failures.}
  \label{tab:yang_comparison}
  \begin{tabular}{@{}>{\raggedright\arraybackslash}p{0.24\linewidth}
                  >{\raggedright\arraybackslash}p{0.33\linewidth}
                  >{\raggedright\arraybackslash}p{0.37\linewidth}@{}}
    \toprule
    & \textbf{\citet{yang:2025numerical}} & \textbf{GAP} \\
    \midrule
    Controlled change & Numerical values in a fixed GSM8K template &
      Identifier realisation, or mathematical setting \\
    Diagnostic split & Calculation vs.\ reasoning error in the output &
      Role-binding failure vs.\ proof-plan-transfer failure \\
    When the split is defined & After observing an incorrect output &
      Before evaluation, by the transformation family \\
    Evaluation scope & 250 GSM8K sources, 6 open models &
      1{,}051 Putnam sources, 18 commercial and open models \\
    \bottomrule
  \end{tabular}
\end{table}

The two decompositions are complementary. Yang et al.'s broad
``reasoning error'' category does not separate a failure caused by
changed identifier names from a failure to carry a proof plan into a
different mathematical setting; conversely, GAP does not classify the
type of error in an incorrect answer. Yang et al.\ label the error
after the fact, whereas GAP fixes in advance which invariance is
tested, through the two transformation families of
\S\ref{sec:two-families}.

%% file: sections/appendix/m_recommendations.tex
\section{Extended recommendations}
\label{app:recommendations}

For \emph{evaluation}, aggregate accuracy or a single robustness index
over mixed transformation types conflates the two failure modes. At
minimum, surface and kernel drops should be reported separately, and
where feasible paired structural-overlap and prefix-restoration
rebound rates should accompany accuracy, so that readers can
distinguish representation-level fragility from plan-level fragility
without re-running the models.

For \emph{model development}, the dichotomy suggests distinct levers
for each failure mode. On the surface side, enforcing self-consistency
of reasoning chains --- so that a model's trajectory remains stable
under identifier permutation --- would directly target the
representation-level fragility documented in \S\ref{sec:dissociation};
training-time consistency regularisation and inference-time
self-verification are both plausible vehicles. On the plan side, the
present findings suggest three further directions. First, supervision
that explicitly pairs different surface and kernel forms of the same
proof plan, so that plan invariance becomes a training target rather
than an incidental side-effect of large-scale exposure. Second,
inference-time retrieval over a structurally indexed bank of solved
instances, so that a candidate plan is conditioned on plans known to
transfer rather than on surface-similar text. Third, decoding
strategies that commit to a high-level plan sketch before
instantiating it, exposing plan choice as an auditable intermediate
rather than leaving it implicit in the trajectory. Direct treatment
of plan-level fragility, rather than another round of surface-level
robustification, is the route we believe most likely to convert
further capability gains into durable mathematical reasoning.